\documentclass[10pt,journal,compsoc]{IEEEtran}

\usepackage{multirow}
\usepackage[table]{xcolor}
\usepackage{algorithm} 
\usepackage{amssymb}
\usepackage{amsfonts}
\usepackage{multicol}
\usepackage{graphicx}
\usepackage{booktabs}
\usepackage[T1]{fontenc}
\usepackage[breaklinks=true,bookmarks=true,pagebackref=false,colorlinks,linkcolor=blue,anchorcolor=red,citecolor=blue,urlcolor=blue]{hyperref}
\usepackage[numbers,sort]{natbib}

\usepackage{comment}
\usepackage{subcaption}
\usepackage{amsmath}
\usepackage{algorithmic}
\usepackage{array}
\usepackage[switch]{lineno}
\usepackage{longtable}
\usepackage{xspace}
\usepackage{url}
\usepackage{overpic}
\usepackage{ragged2e}
\usepackage{framed}
\usepackage{enumitem}

\makeatletter
\DeclareRobustCommand\onedot{\futurelet\@let@token\@onedot}
\def\@onedot{\ifx\@let@token.\else.\null\fi\xspace}
\def\eg{e.g\onedot} 
\def\ie{i.e\onedot} 
 
\def\etc{etc\onedot}

\def\etal{et~al\onedot}

\makeatother

\definecolor{darkgreen}{rgb}{0,0.7,0}
\definecolor{darkblue}{RGB}{31,119,180}
\definecolor{darkred}{RGB}{214,39,40}
\definecolor{mediumgray}{rgb}{0.5,0.5,0.5}
\definecolor{mediumteal}{rgb}{0,0.5,0.5}

\definecolor{ellisred}{rgb}{0.87,0.44,0.38} %
\definecolor{ellisgreen}{rgb}{0.69,0.90,0.52} %
\definecolor{elliscyan}{rgb}{0.29,0.77,0.74} %
\definecolor{ellisorange}{rgb}{0.89,0.55,0.28} %
\definecolor{ellisblue}{rgb}{0.41,0.61,0.86} %

\newcommand{\mrev}[1]{{#1}}

\begin{document}
\title{
End-to-end Autonomous Driving:\\Challenges and Frontiers
}

\author{
    Li Chen,
    Penghao Wu,
    Kashyap Chitta,
    Bernhard Jaeger,
    Andreas Geiger 
    and Hongyang Li

\IEEEcompsocitemizethanks{
\IEEEcompsocthanksitem L. Chen and H. Li are with OpenDriveLab, Shanghai AI Lab, Shanghai, China, and the University of Hong Kong, Hong Kong, China. P. Wu is with OpenDriveLab, Shanghai AI Lab, Shanghai, China.
K. Chitta, B. Jaeger and A. Geiger are with University of Tübingen and Tübingen AI Center, Germany. 
Primary contact: \texttt{lihongyang@pjlab.org.cn}

}%
}

\IEEEtitleabstractindextext{%
\begin{abstract}
\justifying The autonomous driving community has witnessed a rapid growth in approaches that embrace an end-to-end algorithm framework, utilizing raw sensor input to generate vehicle motion plans, instead of concentrating on individual tasks such as detection and motion prediction. End-to-end systems, in comparison to modular pipelines, benefit from joint feature optimization for perception and planning. This field has flourished due to the availability of large-scale datasets, closed-loop evaluation, and the increasing need for autonomous driving algorithms to perform effectively in challenging scenarios. In this survey, we provide a comprehensive analysis of more than 270 papers, covering the motivation, roadmap, methodology, challenges, and future trends in end-to-end autonomous driving. We delve into several critical challenges, including multi-modality, interpretability, causal confusion, robustness, and world models, amongst others. Additionally, we discuss current advancements in foundation models and visual pre-training, as well as how to incorporate these techniques within the end-to-end driving framework. 
We maintain an active repository that contains up-to-date 
literature and open-source projects at 
\texttt{\url{https://github.com/OpenDriveLab/End-to-end-Autonomous-Driving}}.
\end{abstract}

}

\maketitle

\IEEEdisplaynontitleabstractindextext
\IEEEpeerreviewmaketitle

\IEEEraisesectionheading{
\section{Introduction}\label{sec:introduction}
}

\IEEEPARstart{C}{onventional} autonomous driving systems adopt a modular design strategy, wherein each functionality, such as perception, prediction, and planning, is individually developed and integrated into onboard vehicles. The planning or control module, responsible for generating steering and acceleration outputs, plays a crucial role in determining the driving experience.
The most common approach for planning in modular pipelines involves using sophisticated rule-based designs, which are often ineffective in addressing the vast number of situations that occur on road. Therefore, there is a growing trend to leverage large-scale data and to use learning-based planning as a viable alternative.

\begin{figure*}[t]
    \centering
    \includegraphics[width=.98\linewidth]{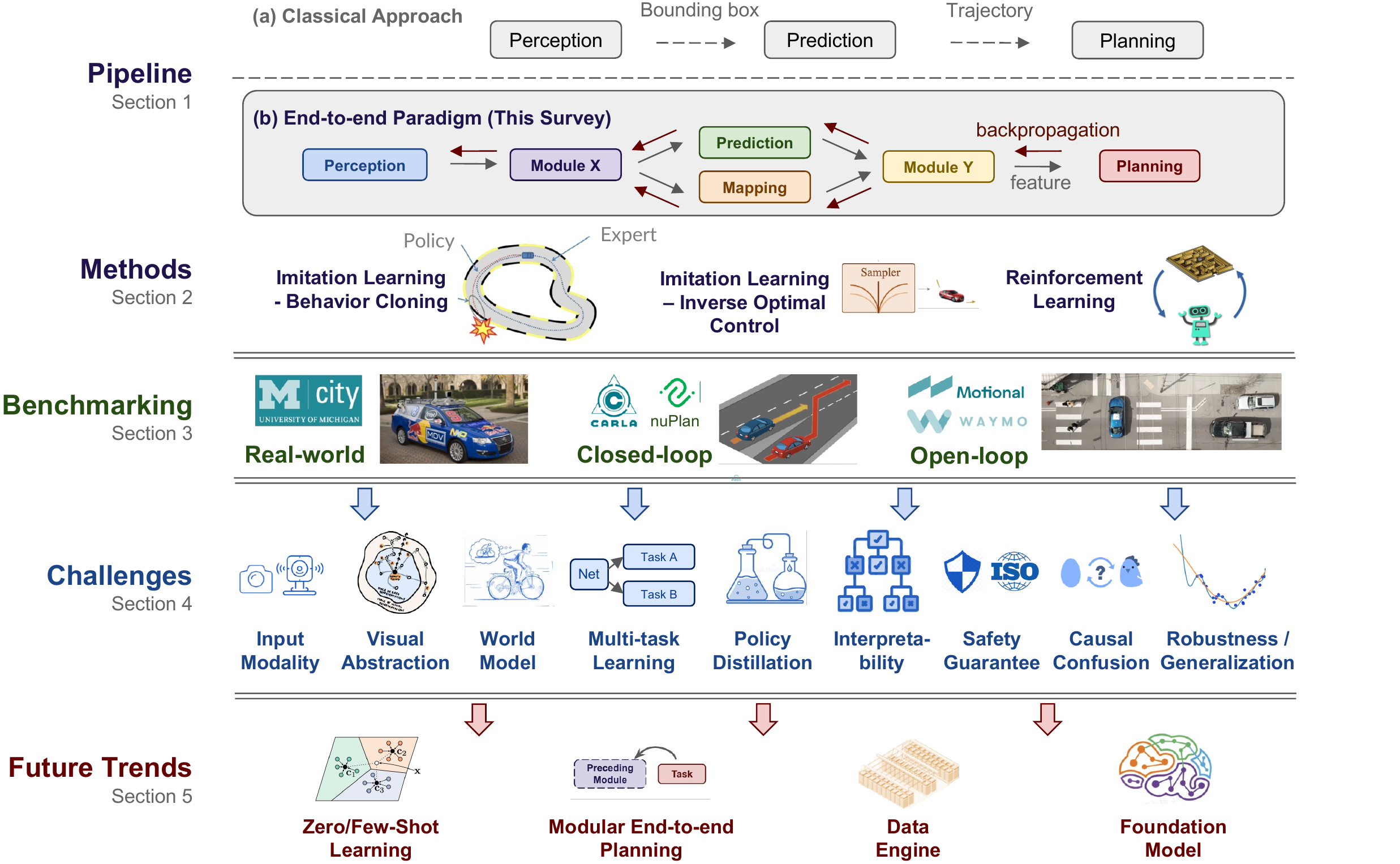}
    \caption{\mrev{\textbf{Survey at A Glance.}} 
    \textbf{(a) Pipeline and Methods.} We define end-to-end autonomous driving as a learning-based algorithm framework with raw sensor input and planning/control output. We deepdive into 270+ papers and categorize into imitation learning (IL) and reinforcement learning (RL).
    \textbf{(b) Benchmarking.} We group popular benchmarks into closed-loop and open-loop evaluation, respectively. We cover various aspects of closed-loop simulation and the limitations of open-loop evaluation for this problem.
    \textbf{(c) Challenges.} %
    This is the main section of our work. We list key challenges from a wide range of topics and extensively analyze why these concerns are crucial. \mrev{Promising resolutions to these challenges are covered as well.} 
    \textbf{(d) Future Trends.} We discuss how end-to-end paradigm could benefit by aid of the rapid development of foundation models, visual pre-training, \textit{\etc}
    Partial photos by courtesy of online resources.
    }
    \label{fig:overview}
\end{figure*}

We define end-to-end autonomous driving systems as fully differentiable programs that take raw sensor data as input and produce a plan and/or low-level control actions as output. Fig.~\ref{fig:overview} (a)-(b) illustrates the difference between the classical and end-to-end formulation.
The conventional approach feeds the output of each component, such as bounding boxes and vehicle trajectories, directly into subsequent units (dashed arrows). In contrast, the end-to-end paradigm propagates feature representations across components (gray solid arrow). The optimized function is set to be, for example, the planning performance, and the loss is minimized via back-propagation (red arrow). Tasks are jointly and globally optimized in this process.

In this survey, we conduct an extensive review of this emerging topic. Fig.~\ref{fig:overview} provides an overview of our work. We begin by discussing the motivation and roadmap for end-to-end autonomous driving systems. End-to-end approaches can be broadly classified into imitation and reinforcement learning, and we give a brief review of these methodologies. We cover datasets and benchmarks for both closed and open-loop evaluation.
We summarize a series of critical challenges, including interpretability, generalization, world models, causal confusion, \textit{etc}. We conclude by discussing future trends that we think should be embraced by the community to incorporate the latest developments from data engines, and large foundation models, amongst others.
\mrev{
Note that this review
is mainly orchestrated from a theoretical perspective. Engineering efforts such as version control, unit testing, data servers, data cleaning, software-hardware co-design, \textit{\etc}, play crucial roles in deploying the end-to-end technology. Publicly available information regarding the latest practices on these topics is limited. We invite the community towards more openness 
in future discussions.
}

\subsection{Motivation of an End-to-end System}
\label{sec:motivation}

In the classical pipeline, each model serves a standalone component and corresponds to a specific task (\textit{\eg}, traffic light detection). Such a design is beneficial in terms of interpretability and ease of debugging. However, since the optimization objectives across modules are different, with detection 
pursuing mean average precision (mAP) while planning aiming for driving safety and comfort, the entire system may not be aligned with a unified target, \textit{\ie}, the ultimate planning/control task. Errors from each module, as the sequential procedure proceeds, could be compounded and result in an information loss.
Moreover, \mrev{compared to one end-to-end neural network,} the multi-task, multi-model deployment \mrev{which involves multiple 
encoders and message transmission systems,} may increase the computational burden and potentially lead to sub-optimal use of compute.

In contrast to its classical counterpart, an end-to-end autonomous system offers several advantages. 
(a) The most apparent merit is its simplicity in combining perception, prediction, and planning into a single model that can be jointly trained. 
(b) The whole system, including its intermediate representations, is optimized towards the ultimate task. 
(c) Shared backbones increase computational efficiency.
(d) Data-driven optimization has the potential to 
improve the system by simply scaling training resources.

Note that the end-to-end paradigm does not necessarily indicate one black box with only planning/control outputs. It could have intermediate representations and outputs (Fig.~\ref{fig:overview} (b)) as in classical approaches. In fact, several state-of-the-art systems \cite{Casas2021CVPR, uniad} propose a modular design but optimize all components together to achieve superior performance.

\subsection{Roadmap}
\label{sec:roadmap}

\begin{figure*}[t]
    \centering
    \begin{overpic}[width=0.96\linewidth]{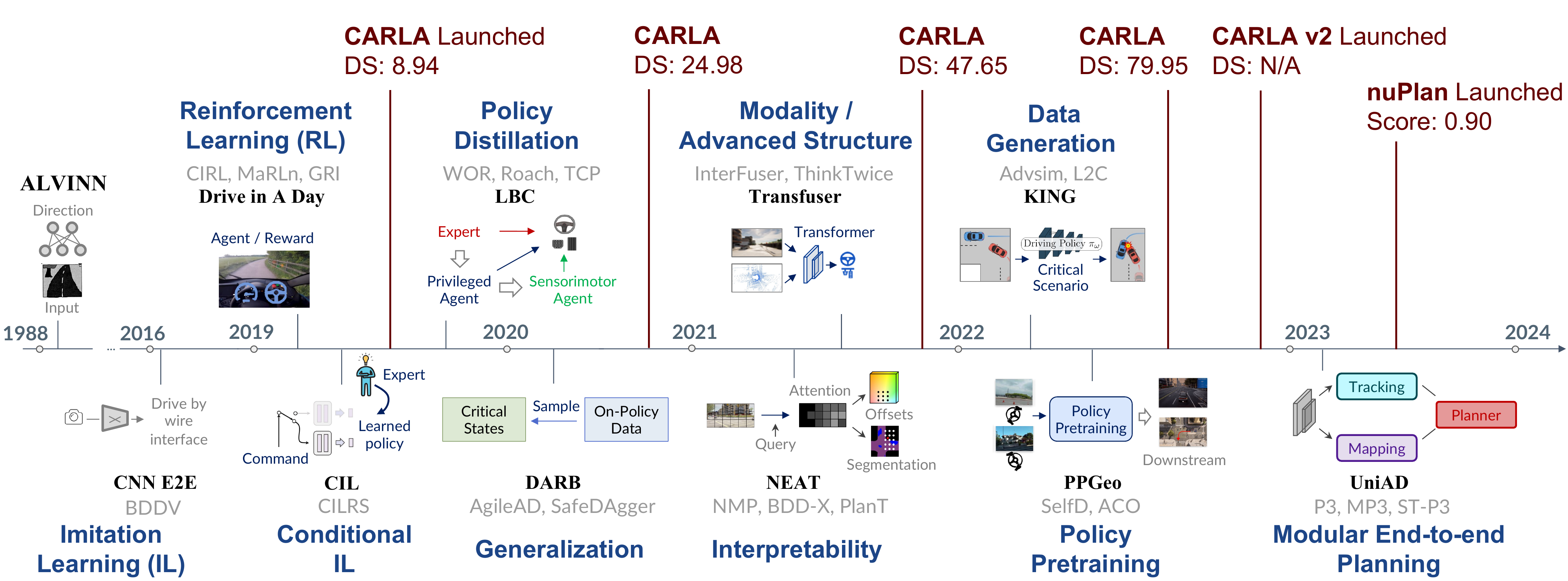}
    
\put(6.7,24.9){\scriptsize~\cite{pomerleau1988alvinn}}

\put(20.5,24.0){\scriptsize~\cite{l2diad}}

\put(34.0,24.0){\scriptsize~\cite{LBC}}

\put(53.5,24.0){\scriptsize~\cite{transfuser2021CVPR}}

\put(68.5,24.0){\scriptsize~\cite{KING}}

\put(12.4,5.8){\scriptsize~\cite{DAVE-2}}

\put(22.9, 5.8){\scriptsize~\cite{CIL}}

\put(36.8, 5.8){\scriptsize~\cite{DARB}}

\put(52.1, 5.8){\scriptsize~\cite{NEAT}}

\put(71.3, 5.8){\scriptsize~\cite{ppgeo}}

\put(89.7, 5.8){\scriptsize~\cite{uniad}}
\end{overpic}
\vspace{-4pt}
    \caption{\textbf{Roadmap of End-to-end Autonomous Driving.} 
    We present the key milestones chronologically, grouping similar works under the same theme. The representative or first work is shown in bold with an illustration, while the date of the rest of the literature in the same theme may vary. We also display the score for each year's top entry in the CARLA leaderboard \cite{carlaleaderboard} (DS, ranging from 0 to 100) and the recent nuPlan challenge \cite{nuplan} (Score ranging from 0 to 1).
    }
    \label{fig:roadmap}
\end{figure*}
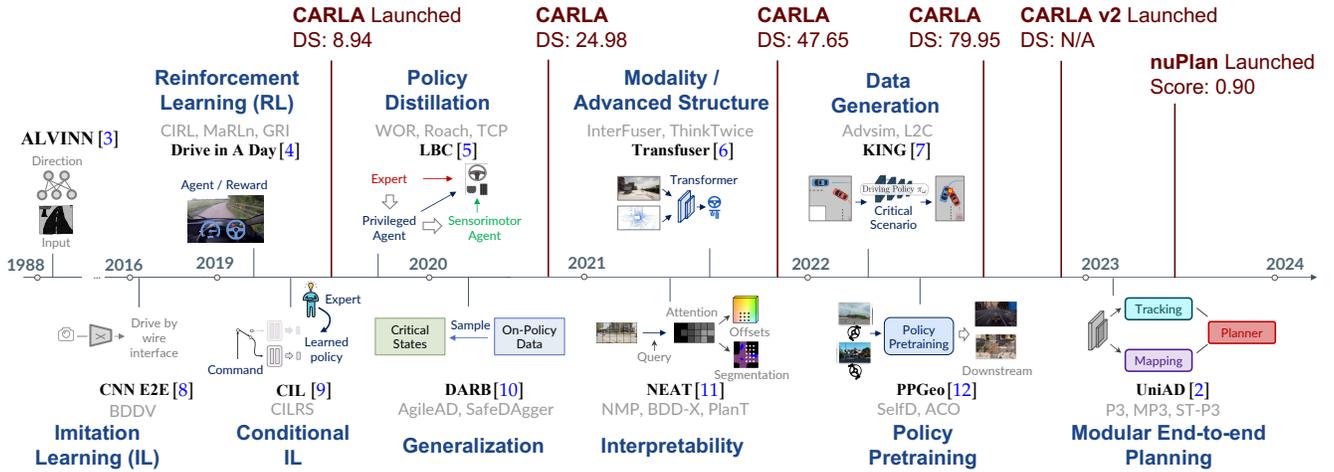

Fig.~\ref{fig:roadmap} depicts a chronological roadmap of critical achievements in end-to-end autonomous driving, where each part indicates an essential paradigm shift or performance boost.
The history of end-to-end autonomous driving dates back to 1988 with ALVINN~\cite{pomerleau1988alvinn}, where the input was two ``retinas`` from a camera and a laser range finder, and a simple neural network generated steering output. NVIDIA designed a prototype end-to-end CNN system,
which reestablished this idea in the new era of GPU computing~\cite{DAVE-2}.
Notable progress has been achieved with the development of deep neural networks, both in imitation learning \cite{UB-CIL, CILRS} and reinforcement learning \cite{l2diad, liang2018cirl, MaRLn, GRI}. 
The policy distillation paradigm proposed in LBC \cite{LBC} and related approaches \cite{WoR, roach, Wu2022NeurIPS, zhang2023coaching} has significantly improved closed-loop performance by mimicking a well-behaved expert. 
To enhance generalization ability due to the discrepancy between the expert and learned policy, several papers \cite{DARB, AgileAD, SafeDAgger} have proposed aggregating on-policy data \cite{DAgger} during training.

A significant turning point occurred around 2021.
With diverse sensor configurations available within a reasonable computational budget, attention was focused on incorporating more modalities and advanced architectures (\textit{\eg}, Transformers \cite{transformer}) to capture global context and representative features, as in TransFuser \cite{transfuser2021CVPR, transfuser2022PAMI} and many variants \cite{Shao2022CORL, jia2023thinktwice, Jaeger2023ICCV}. Combined with more insights about the simulation environment, these advanced designs resulted in a substantial performance boost on the CARLA benchmark~\cite{carlaleaderboard}.
To improve the interpretability and safety of autonomous systems, approaches~\cite{NEAT, Zeng2019CVPR, BDD-X} 
explicitly involve various auxiliary modules to better supervise the learning process or utilize attention visualization.  
Recent works prioritize generating safety-critical data \cite{KING, Advsim, Learningtocollide}, pre-training a foundation model or backbone curated for policy learning \cite{ACO, ppgeo, selfD}, and advocating a modular end-to-end planning philosophy \cite{uniad, hu2022stp3, P3, Casas2021CVPR}. Meanwhile, the new and challenging CARLA v2 \cite{carlaleaderboard} and nuPlan \cite{nuplan} benchmarks have been introduced to facilitate research into this area.

\subsection{Comparison to Related Surveys}
\label{sec:related-survey}

We would like to clarify the difference between our survey and previous related surveys \cite{Janai2017ARXIV, Tampuu2020NNLS, teng2023motion, coelho2022review, Ly2020TIV, LeMero2022TITS, zheng2022imitation, Zhu2021TITS, Kiran2022TITS}. 
Some prior surveys \cite{Janai2017ARXIV, Tampuu2020NNLS, coelho2022review, teng2023motion} cover content similar to ours in the sense of an end-to-end system. However, they do not cover new benchmarks and approaches that arose with the significant recent transition in the field, and place a minor emphasis on frontiers and challenges. 
The others focus on specific topics in this domain, such as imitation learning \cite{Ly2020TIV, LeMero2022TITS, zheng2022imitation} or reinforcement learning \cite{Zhu2021TITS, Kiran2022TITS}.
In contrast, our survey provides up-to-date information on the latest developments
in this field, covering a wide span of topics and providing in-depth discussions of critical challenges.

\subsection{Contributions}
\label{sec:contribution}
To summarize, this survey has three key contributions:

\noindent \textbf{(a)} We provide a comprehensive analysis of end-to-end autonomous driving for the first time, including high-level motivation, methodologies, benchmarks, and more. Instead of optimizing a single block, we advocate for a philosophy to design the algorithm framework as a whole, with the ultimate target of achieving safe and comfortable driving.

\noindent \textbf{(b)} We extensively investigate the critical challenges that concurrent approaches face. Out of the more than 270 papers surveyed, we summarize major aspects and provide in-depth analysis, including topics on generalizability, language-guided learning, causal confusion, \textit{etc}.

\noindent \textbf{(c)} We cover the broader impact of how to embrace large foundation models and data engines. We believe that this line of research and the large scale of high-quality data it provides could significantly advance this field. To facilitate future research, we maintain an active repository updated with new literature and open-source projects.

\begin{figure*}[t]
    \centering
    \includegraphics[width=.98\linewidth]{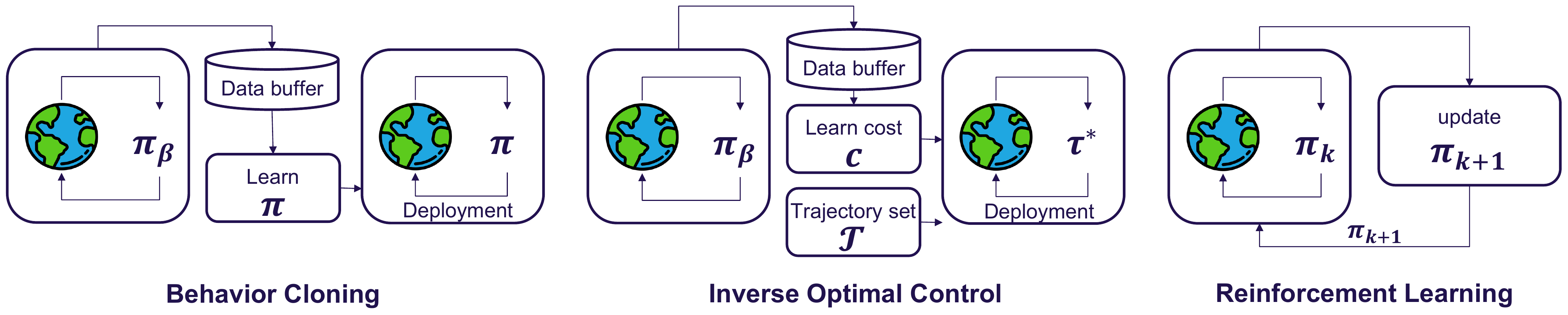}
    \vspace{-3pt}
    \caption{\textbf{Overview of methods in end-to-end autonomous driving.} We illustrate three popular paradigms, including two imitation learning frameworks (behavior cloning and inverse optimal control), as well as online reinforcement learning. 
    }
    \label{fig:method}
\end{figure*}

\section{Methods}
\label{sec:method}

This section reviews fundamental principles behind most existing end-to-end self-driving approaches.
Sec.~\ref{sec:IL} discusses methods using imitation learning and provides details on the two most popular sub-categories, namely behavior cloning and inverse optimal control. Sec.~\ref{sec:RL} summarizes methods that follow the reinforcement learning paradigm. 

\subsection{Imitation Learning}
\label{sec:IL}

Imitation learning (IL), also referred to as learning from demonstrations, trains an agent to learn the 
policy by imitating the behavior of an expert. IL requires a dataset $D=\{\xi_i\}$ containing trajectories collected under the expert's policy $\pi_\beta$, where each trajectory is a sequence of state-action pairs.
The goal of IL is to learn an agent policy $\pi$ that matches $\pi_\beta$. 

\mrev{The policy $\pi$ can output planned trajectories or control signals. Early works usually adopt control outputs, due to the ease of collection. However, predicting controls at different steps could lead to discontinuous maneuvers and the network inherently specializes to the vehicle dynamics which hinders generalization to other vehicles.
Another genre of works predicts waypoints. It considers a relatively longer time horizon. Meanwhile, converting trajectories for vehicles to track into control signals needs additional controllers, which is non-trivial and involves vehicle models and control algorithms.
Since no clear performance gap has been observed between these two paradigms, we do not differentiate them explicitly in this survey. An interesting and more in-depth discussion 
can be found in \cite{Wu2022NeurIPS}.
}

One widely used category of IL is behavior cloning (BC) \cite{BC}, which reduces the problem to supervised learning. Inverse Optimal Control (IOC), also known as Inverse Reinforcement Learning (IRL) \cite{IRL-MEIRL} is another type of IL method that utilizes expert demonstrations to learn a reward function. We elaborate on these two categories below.

\subsubsection{Behavior Cloning}
\label{sec:BC}

In BC, matching the agent's policy with the expert's is accomplished by minimizing planning loss as supervised learning over the collected dataset: $\mathop{\mathbb{E}}_{(s,a)} \ell(\pi_\theta(s), a)$. Here, $\ell(\pi_\theta(s), a)$ represents a loss function that measures the distance between the agent action and the expert action. 

Early applications of BC for driving \cite{pomerleau1988alvinn, DAVE, DAVE-2} utilized an end-to-end neural network to generate control signals from camera inputs. Further enhancements, such as multi-sensor inputs \cite{transfuser2021CVPR, chen2022lav}, auxiliary tasks \cite{CILRS, transfuser2022PAMI}, and improved expert design \cite{roach}, have been proposed to enable BC-based end-to-end driving models to handle challenging urban scenarios.

BC is advantageous due to its simplicity and efficiency, as it does not require hand-crafted reward design, which is crucial for RL. However, there are some common issues.
During training, it treats each state as independently and identically distributed, resulting in an important problem known as covariate shift. For general IL, several on-policy methods have been proposed to address this issue \cite{activeIL, SMILe, ross2014reinforcement, DAgger}. In the context of end-to-end autonomous driving, DAgger \cite{DAgger} has been adopted in \cite{DARB, LBC, MetaDAgger, SafeDAgger}. Another common problem with BC is causal confusion, where the imitator exploits and relies on false correlations between certain input components and output signals. This issue has been discussed in the context of end-to-end autonomous driving in \cite{Wen2020NEURIPS, Wen2021ICML, Park2021NEURIPS, Wen2022ICML}. These two challenging problems 
are further discussed in Sec.~\ref{sec: challenge generalizability} and Sec.~\ref{sec:causal_confusion}, respectively.

\subsubsection{Inverse Optimal Control}
\label{sec:IOC}

Traditional IOC algorithms learn an unknown reward function $R(s, a)$ 
from expert demonstrations, where the expert's reward function can be represented as a linear combination of features \cite{IRL-brown2019extrapolating, IRL-MEIRL, IRL-POIRL, IRL-reddy2019sqil, IRL-SILP}. However, in continuous, high-dimensional autonomous driving scenarios, the definition of the reward is implicit and difficult to optimize. 

Generative adversarial imitation learning~\cite{GAIL, InfoGAIL, MixGAIL} is a specialized approach in IOC that designs the reward function as an adversarial objective to distinguish the expert and learned policies, similar to the concept of generative adversarial networks~\cite{GAN}.
Recently, several works propose optimizing a cost volume or cost function with auxiliary perceptual tasks. Since a cost is an alternative representation of the reward, we classify these methods as belonging to the IOC domain. We define the cost learning framework as follows: end-to-end approaches learn a reasonable cost $c(\cdot)$
and use 
algorithmic trajectory samplers to select the trajectory $\tau^*$ with the minimum cost, as illustrated in Fig.~\ref{fig:method}. 

Regarding cost design, 
it has representations including a learned cost volume in a bird's-eye-view (BEV)~\cite{Zeng2019CVPR}, joint energy calculated from other agents' future motion~\cite{INMP}, or a set of probabilistic semantic occupancy or freespace layers~\cite{P3, FF, EO}. 
On the other hand, trajectories are typically sampled from a fixed expert trajectory set \cite{Casas2021CVPR, Chen2024VADv2} or processed by parameter sampling with a kinematic model \cite{Zeng2019CVPR, P3, hu2022stp3, FF}. Then, a max-margin loss is adopted as in classic IOC methods to encourage the expert demonstration to have a minimal cost while others have high costs.

Several challenges exist with cost learning approaches. In particular, in order to generate more realistic costs, HD maps, auxiliary perception tasks, and multiple sensors are typically incorporated, which increases the difficulty of learning and constructing datasets for multi-modal multi-task frameworks. 
Nevertheless, the aforementioned cost learning methods significantly enhance the safety and interpretability of decisions (see Sec.~\ref{sec: challenge interpretability}), and we believe that the industry-inspired end-to-end system design is a viable approach for real-world applications.

\subsection{Reinforcement Learning}
\label{sec:RL}

Reinforcement learning (RL) \cite{Sutton1998TNNLS, Jaeger2023ARXIV} is a field of learning by trial and error. The success of deep Q networks (DQN)~\cite{mnih2015nature} in achieving human-level control on the Atari benchmark \cite{bellemare2013arcade} has popularized deep RL. DQN trains a neural network called the critic (or Q network), which takes as input the current state and an action, and predicts the discounted return of that action. 
The policy is then implicitly defined by selecting the action with the highest predicted return. 

RL requires an environment that allows potentially unsafe actions to be executed, to collect novel data (\textit{\eg}, via random actions). Additionally, RL requires significantly more data to train than IL. For this reason, modern RL methods often parallelize data collection across multiple environments \cite{APEX}. Meeting these requirements in the real world presents great challenges. Therefore, almost all papers that use RL in driving have only investigated the technique in simulation. Most use different extensions of DQN. The community has not yet converged on a specific RL algorithm.

RL has successfully learned lane following on a real car on an empty street~\cite{l2diad}. Despite this encouraging result, it must be noted that a similar task was already accomplished by IL three decades prior~\cite{pomerleau1988alvinn}. To date, no report has shown results for end-to-end training with RL that are competitive with IL. 
The reason for this failure likely is that the gradients obtained via RL are insufficient to train deep perception architectures (\textit{\ie}, ResNet) required for driving. Models used in benchmarks like Atari, where RL succeeds, are relatively shallow, consisting of only a few layers~\cite{Bjorck2021NeurIPS}.

RL has been successfully applied in end-to-end driving when combined with supervised learning (SL). Implicit affordances \cite{MaRLn, GRI} pre-train the CNN encoder using SL with tasks like semantic segmentation. In the second stage, this encoder is frozen, and a shallow policy head is trained on the features from the frozen encoder with a modern version of Q-learning~\cite{toromanoff2019ArXiv}. 
RL can also be used to finetune full networks that were pre-trained using IL~\cite{liang2018cirl, Ohn-Bar2020CVPR}. 

RL can also been effectively applied, if the network has access to privileged simulator information.~\cite{Kiran2022TITS, Knox2021AI, Zhang2022ECCV}.
Privileged RL agents can be used for dataset curation. Roach~\cite{roach} trains an RL agent on privileged BEV semantic maps and uses the policy to automatically collect a dataset with which a downstream IL agent is trained. WoR \cite{WoR} employs a Q-function and tabular dynamic programming to generate additional or improved labels for a static dataset.

A challenge in the field is to transfer the findings from simulation to the real world. In RL, the objective is expressed as reward functions, and many algorithms require them to be dense and provide feedback at each environment step. Current works typically use simple objectives, such as progress and collision avoidance.
These simplistic designs potentially encourage risky behaviors~\cite{Knox2021AI}. Devising or learning better reward functions remains an open problem. Another direction would be to develop RL algorithms that can handle sparse rewards, enabling the optimization of relevant metrics directly. RL can be effectively combined with world models~\cite{dreamer, dreamerv2, Ha2018NEURIPS}, though this presents specific challenges (See Sec.~\ref{sec: challenge world model}). Current RL solutions for driving rely heavily on low-dimensional representations of the scene, and this issue is further discussed in Sec.~\ref{sec: challenge representation learning - pretrain}.

\section{Benchmarking}
\label{sec:benchmarking}

Autonomous driving systems require a comprehensive evaluation to ensure safety. Researchers must benchmark these systems using appropriate datasets, simulators, metrics, and hardware to accomplish this. This section delineates three approaches for benchmarking end-to-end autonomous driving systems: \mrev{(1) real-world evaluation, (2) online or closed-loop evaluation in simulation, and (3) offline or open-loop evaluation on driving datasets}. We focus on the scalable and principled online simulation setting and summarize real-world and offline assessments for completeness.

\subsection{\mrev{Real-world Evaluation}}
\label{sec: Real-world Evaluation}

\mrev{Early efforts on benchmarking self-driving involved real-world evaluation. Notably, DARPA initiated a series of races.
The first event offered \$1M in prize money for autonomously navigating a 240km route through the Mojave desert, which no team achieved~\cite{buehler20072005}. 
The final series event, called the DARPA Urban Challenge, required vehicles to navigate a 96km mock-up town course, adhering to traffic laws and avoiding obstacles~\cite{buehler2009darpa}. These races fostered important developments in self-driving, such as LiDAR sensors. Following this spirit, the University of Michigan established MCity~\cite{mcity}, a large controlled real-world environment to facilitate testing autonomous vehicles. 
However, such academic ventures have not been widely employed for end-to-end systems due to a lack of data and vehicles.
In contrast, industries with the resources to deploy fleets of driverless vehicles could rely on real-world evaluation to benchmark improvements in their algorithms. 
}

\subsection{Online/Closed-loop Simulation}
\label{sec: Online Evaluation}

Conducting tests of self-driving systems in the real world is costly and risky. To address this challenge, simulation is a viable alternative~\cite{wymann2000torcs, martinez2017gtav, Dosovitskiy2017CORL, deepdrive, li2022metadrive, nuplan}. \mrev{Simulators facilitate rapid prototyping and testing, enable the quick iteration of ideas, and provide low-cost access to diverse scenarios for unit testing. In addition, simulators offer tools for measuring performance accurately.} However, their primary disadvantage is that the results obtained in a simulated environment do not necessarily generalize to the real world (Sec.~\ref{sec: domain-adaptation}).

Closed-loop evaluation involves building a simulated environment that closely mimics a real-world driving environment. The evaluation entails deploying the driving system in simulation and measuring its performance. The system has to navigate safely through traffic while progressing toward a designated goal location. There are four main sub-tasks involved in developing such simulators: parameter initialization, traffic simulation, sensor simulation, \mrev{and vehicle dynamics simulation}. We briefly describe these sub-tasks below, followed by a summary of currently available open-source simulators for closed-loop benchmarks.

\subsubsection{Parameter Initialization}
\label{sec:init_sim}

Simulation offers the benefit of a high degree of control over the environment, including weather, maps, 3D assets, and low-level attributes such as the arrangement of objects in a traffic scene. While powerful, the number of these parameters is substantial, resulting in a challenging design problem. Current simulators tackle this in two ways:

\textbf{Procedural Generation:} Traditionally, initial parameters are hand-tuned by 3D artists and engineers~\cite{wymann2000torcs, martinez2017gtav, Dosovitskiy2017CORL, deepdrive}. This limits scalability. 
\mrev{Recently, some of the simulation properties can be sampled from a probabilistic distribution with computer algorithms, which we refer to as procedural generation~\cite{Hendrikx2013procedural}.}
Procedural generation algorithms combine rules, heuristics, and randomization to create diverse road networks, traffic patterns, lighting conditions, and object placements~\cite{fremont2019scenic, hauer2019did}. Due to its efficiency compared to fully manual design, it has become one of the most commonly used methods of initialization for video games and simulations. Nevertheless, the process still needs pre-defined parameters and algorithms to control generation reliability, which is time-consuming and requires a lot of expertise. 

\textbf{Data-Driven:} Data-driven approaches for simulation initialization aim to learn the required parameters. Arguably, the simplest way is to sample from real-world driving logs~\cite{li2022metadrive, nuplan}, where parameters such as road maps or traffic patterns are directly extracted from pre-recorded datasets. The advantage of log sampling is its ability to capture the natural variability present in real-world data, leading to more realistic simulation scenarios. However, it may not encompass rare situations that are critical for testing the robustness of autonomous driving systems. The initial parameters can be optimized to increase the representation of such scenarios \cite{KING, Advsim, Learningtocollide}. Another advanced data-driven approach to initialization is generative modeling, where machine learning algorithms are utilized to learn the underlying structure and distributions of real-world data. They can then generate novel scenarios that resemble the real world but were not included in the original data \cite{tan2021scenegen, Bergamini2021ICRA, Feng2023ICRA, Chitta2024ARXIV}.

\subsubsection{Traffic Simulation}
\label{sec:traffic_sim}

Traffic simulation involves generating and positioning virtual entities in the environment with realistic motion~\cite{Bergamini2021ICRA, Suo2021CVPR}. These entities often include vehicles (such as cars, motorcycles, bicycles, \textit{\etc}) and pedestrians. Traffic simulators must account for the effects of speed, acceleration, braking, obstructions, and the behavior of other entities. Moreover, traffic light states must be periodically updated to simulate realistic city driving. There are two popular approaches for traffic simulation, which we describe below.

\textbf{Rule-Based:}  Rule-based traffic simulators use pre-defined rules to generate the motion of traffic entities. 
The most prominent implementation of this concept is the Intelligent Driver Model (IDM)~\cite{treiber2000IDM}. IDM is a car-following model that computes acceleration for each vehicle based on its current speed, the speed of the leading vehicle, and a desired safety distance. 
Although widely used and straightforward, this approach may be inadequate to simulate realistic motion and complex interactions in urban environments.

\textbf{Data-Driven:} Realistic human traffic behavior is highly interactive and complex, including lane changing, merging, sudden stopping, \textit{etc}. To model such behavior, data-driven traffic simulation utilizes data collected from real-world driving. These models can capture more nuanced, realistic behavior but require significant amounts of labeled data for training. A wide variety of learning-based techniques have been proposed for this task~\cite{Bergamini2021ICRA, Suo2021CVPR, Zhong2023ICRA, Xu2023ICRA, Zhang2023ICRA, Feng2023ICRA}.

\subsubsection{Sensor Simulation}
\label{sec:sensor_sim}

Sensor simulation is crucial for evaluating end-to-end self-driving systems. This involves generating simulated raw sensor data, such as camera images or LiDAR scans that the driving system would receive from different viewpoints in the simulator~\cite{Manivasagam2020CVPR, Chen2021CVPR, Yang2023CVPR}. This process needs to take into account noise and occlusions to realistically assess the autonomous system. There are two main branches of ideas concerning sensor simulation, as described below.

\textbf{Graphics-Based:} Recent computer graphics simulators use 3D models of the environment, along with traffic entity models, to generate sensor data via approximations of physical rendering processes in the sensors~\cite{martinez2017gtav, Dosovitskiy2017CORL}. For example, this can involve occlusions, shadows, and reflections present in real-world environments while simulating camera images. However, the realism of graphics-based simulation is often subpar or comes at the cost of heavy computation, making parallelization non-trivial~\cite{petrenko2021megaverse}. It is closely tied to the quality of the 3D models and the approximations used in modeling the sensors. A comprehensive survey of graphics-based rendering for driving data is provided in~\cite{song2023synthetic}.
 
\textbf{Data-Driven:} Data-driven sensor simulation leverages real-world sensor data to create the simulation where both the ego vehicle and background traffic may move differently from the way they did in recordings~\cite{vista, vista2, vista2ma}. Popular methods are Neural Radiance Fields (NeRF) \cite{mildenhall2020nerf} and 3D Gaussian Splatting~\cite{kerbl20233d}, which can generate novel views of a scene by learning an implicit representation of the scene's geometry and appearance. These methods can produce more realistic sensor data visually than graphics-based approaches, but they have limitations such as high rendering times or requiring independent training for each scene being reconstructed~\cite{tancik2022block, Turki2022CVPR, Kundu2022CVPR, Yang2023ARXIV, Yang2023CVPR}. Another approach to data-driven sensor simulation is domain adaptation, which aims to minimize the gap between real and graphics-based simulated sensor data \cite{Richter2022PAMI}. Deep learning techniques such as GANs can be employed to improve realism (Sec.~\ref{sec: domain-adaptation}).

\subsubsection{\mrev{Vehicle Dynamics Simulation}}
\label{sec:vehicle_sim}

\mrev{The final aspect of driving simulation pertains to ensuring that the simulated vehicle adheres to physically plausible motion. Most existing publicly available simulators use highly simplified vehicle models, such as the unicycle model~\cite{schoonwinkel1988design} or the bicycle model~\cite{Polack2017IV}. However, in order to facilitate seamless transfer of algorithms from simulation to the real world, it is essential to incorporate more accurate physical modeling of vehicle dynamics. For instance, CARLA adopts a multi-body system approach, representing a vehicle as a collection of sprung masses on four wheels. For a comprehensive review, please refer to~\cite{rajamani2011vehicle}.}

\begin{table}[t]
    \setlength{\tabcolsep}{2pt}
    \centering
    \begin{tabular}{l p{0.8\columnwidth}}
    \toprule
         \textbf{Simulator} & \textbf{Benchmarks} \\
    \midrule
        CARLA & CoRL~\cite{Dosovitskiy2017CORL}, noCrash~\cite{Codevilla2018ECCV},
        Town05~\cite{transfuser2021CVPR}, LAV~\cite{chen2022lav},
        Roach~\cite{roach},
        Longest6~\cite{transfuser2022PAMI}, Leaderboard v1 and v2~\cite{carlaleaderboard}\\
        \midrule
        nuPlan & NAVSIM~\cite{Dauner2024NAVSIM}, Val14~\cite{Dauner2023ARXIV}, Leaderboard~\cite{nuplan}\\
    \bottomrule
    \end{tabular}
    \caption{\textbf{Open-source Simulators} with active benchmarks for closed-loop evaluation of autonomous driving.}
    \label{tab:simulator}
\end{table}

\subsubsection{Benchmarks}
\label{sec:benchmarks}

We give a succinct overview of end-to-end driving benchmarks available up to date in Table~\ref{tab:simulator}. In 2019, the original benchmark released with CARLA~\cite{Dosovitskiy2017CORL} was solved with near-perfect scores~\cite{LBC}. The subsequent NoCrash benchmark \cite{Codevilla2018ECCV} involves training on a single CARLA town under specific weather conditions and testing generalization to another town and set of weathers. Instead of a single town, the Town05 benchmark~\cite{transfuser2021CVPR} involves training on all available towns while withholding Town05 for testing. Similarly, the LAV benchmark trains on all towns except Town02 and Town05, which are both reserved for testing.
Roach~\cite{roach} uses a setting with 3 test towns, albeit all seen during training, and without the safety-critical scenarios in Town05 and LAV. Finally, the Longest6 benchmark~\cite{transfuser2022PAMI} uses 6 test towns. Two online servers, the leaderboard (v1 and v2)~\cite{carlaleaderboard}, ensure fair comparisons by keeping evaluation routes confidential. Leaderboard v2 is highly challenging due to the long route length (over 8km on average, as opposed to 1-2km on v1) and a wide variety of new traffic scenarios. %

The nuPlan simulator is currently accessible for evaluating end-to-end systems via the NAVSIM project~\cite{Dauner2024NAVSIM}. Further, there are two benchmarks on which agents input maps and object properties via the data-driven parameter initialization for nuPlan (Sec.~\ref{sec:init_sim}). Val14, proposed in~\citep{Dauner2023ARXIV}, uses a validation split of nuPlan. The leaderboard, a submission server with the private test set, was used in the 2023 nuPlan challenge, but it is no longer public for submissions.

\subsection{Offline/Open-loop Evaluation}
\label{sec: offline evaluation}

Open-loop evaluation mainly assesses a system’s performance against pre-recorded expert driving behavior. This method requires evaluation datasets that include (1) sensor readings, (2) goal locations, and (3) corresponding future driving trajectories, usually obtained from human drivers. Given sensor inputs and goal locations as inputs, performance is measured by comparing the system’s predicted future trajectory against the trajectory in the driving log. Systems are evaluated based on how closely their trajectory predictions match the human ground truth, as well as auxiliary metrics such as the collision probability with other agents. The advantage of open-loop evaluation is that it is easy to implement \mrev{using} realistic traffic and sensor data, as it does not require a simulator. However, the key disadvantage is that it does not measure performance in the actual test distribution encountered during deployment. During testing, the driving system may deviate from the expert driving corridor, and it is essential to verify the system's ability to recover from such drift (Sec.~\ref{sec: covariate-shift}). Furthermore, the distance between the predicted and the recorded trajectories is not an ideal metric in a multi-modal scenario. For example, in the case of merging into a turning lane, both the options of merging immediately or later could be valid, but open-loop evaluation penalizes the option that was not observed in the data. 
\mrev{Therefore, besides measuring collision probability and prediction errors, a few metrics were proposed to cover more comprehensive aspects such as traffic violations, progress, and driving comfort~\cite{Dauner2023ARXIV}.}

This approach requires comprehensive datasets of trajectories to draw from. The most popular datasets for this purpose include nuScenes~\cite{Caesar2020CVPR}, Argoverse~\cite{wilson2021NEURIPS}, Waymo~\cite{waymo}, and nuPlan~\cite{nuplan}. All of these datasets comprise a large number of real-world driving traversals with varying degrees of difficulty. However, open-loop results do not provide conclusive evidence of improved driving behavior in closed-loop, due to the aforementioned drawbacks~\cite{Codevilla2018ECCV, zhai2023ADMLP, Dauner2023ARXIV, Li2024ego}. Overall, a realistic closed-loop benchmarking, if available and applicable, is recommended in future research.

\section{Challenges}
\label{sec:challenges}

Following each topic illustrated in Fig.~\ref{fig:overview}, we now walk through \mrev{current challenges, related works or potential resolutions, risks, and opportunities}. We start with challenges in handling different input modalities in Sec.~\ref{sec:challenge inputs}, followed by a discussion on visual abstraction for efficient policy learning in Sec.~\ref{sec: challenge representation learning}. Further, we introduce learning paradigms such as world model learning (Sec.~\ref{sec: challenge world model}), multi-task frameworks (Sec.~\ref{sec: challenge MTL}), and policy distillation (Sec.~\ref{sec: challenge distillation}). Finally, we discuss general issues that impede safe and reliable end-to-end autonomous driving, including interpretability in Sec.~\ref{sec: challenge interpretability}, safety guarantees in Sec.~\ref{sec: challenge safety guarantee}, causal confusion in Sec.~\ref{sec:causal_confusion}, and robustness in Sec.~\ref{sec: challenge generalizability}.

\begin{figure}[t]
    \centering
    \includegraphics[width=.95\linewidth]{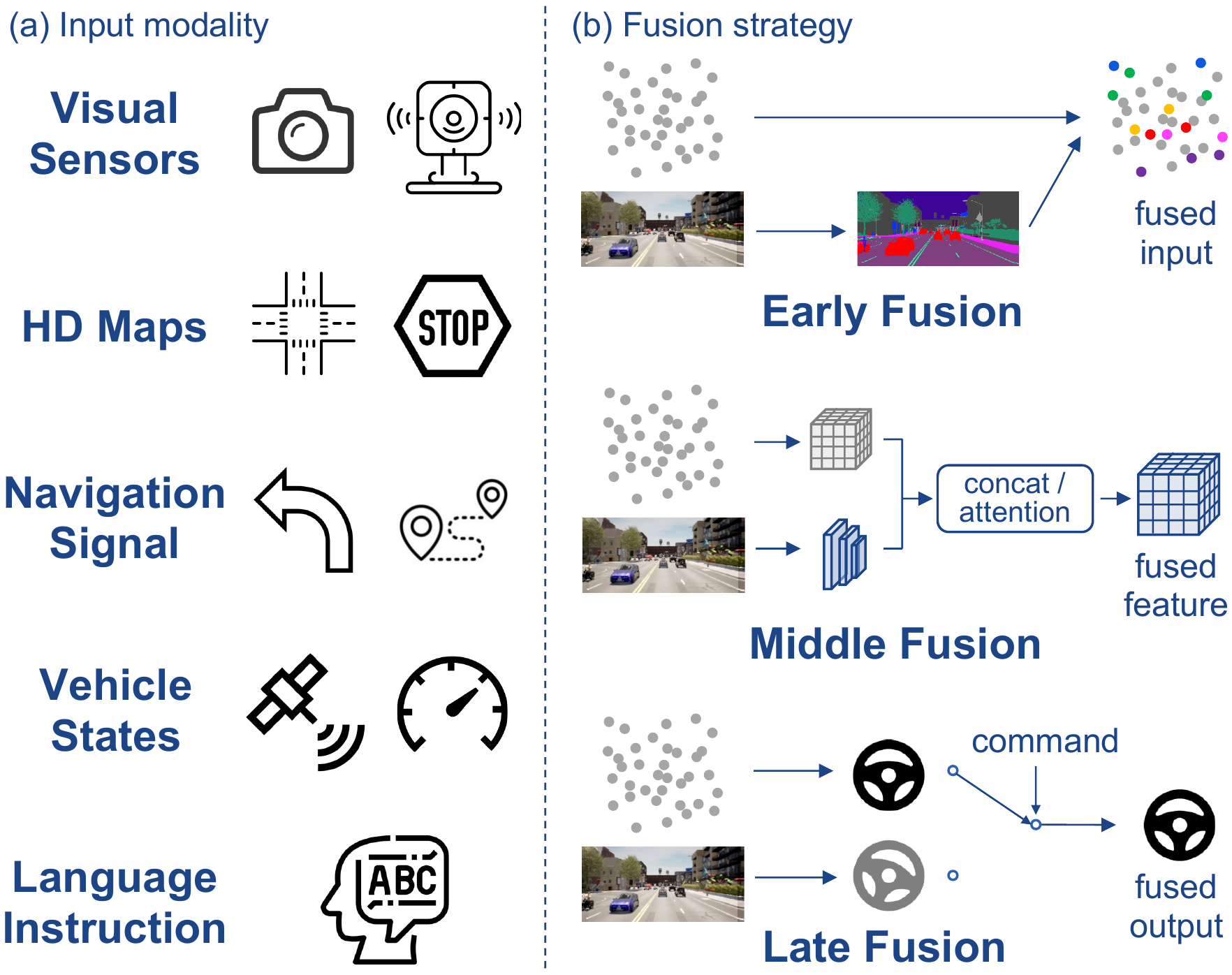}
    \caption{\textbf{Examples of input modality and fusion strategy.} Different modalities have distinct characteristics, leading to the challenge of effective sensor fusion.
    We take point clouds and images as examples to depict various fusion strategies.}
    \label{fig:sensor-inputs}
\end{figure}

\subsection{\mrev{Dilemma over Sensing and Input Modalities}}
\label{sec:challenge inputs}

\subsubsection{\mrev{Sensing and Multi-sensor Fusion}}
\label{sec: challenge multi-sensor fusion}

\textbf{Sensing:} Though early work~\cite{DAVE-2} successfully achieved following
a lane with a monocular camera, this single input modality cannot handle complex scenarios. Therefore, various sensors in Fig.~\ref{fig:sensor-inputs} have been introduced for recent self-driving vehicles. Particularly, RGB images from cameras replicate how humans perceive the world, with abundant semantic details; LiDARs or stereo cameras provide accurate 3D spatial knowledge. 
\mrev{Emerging sensors like mmWave radars and event cameras excel at capturing objects' relative movement.}
Additionally, vehicle states 
from speedometers and IMUs, together with navigation commands, are other lines of input that guide the driving system. 
\mrev{However, various sensors possess distinct perspectives, data distributions, and huge price gaps, thereby posing challenges in effectively designing the sensory layout and fusing them to complement each other for autonomous driving.}

\textbf{Multi-sensor fusion} has predominantly been discussed in perception-related fields, \textit{\eg}, object detection \cite{liang2022bevfusion, liu2022bevfusion}
and semantic segmentation \cite{zhang2015sensor, Meyer2019CVPRWORK}, and is typically categorized into three groups: early, mid, and late fusion. End-to-end autonomous driving algorithms explore similar fusion schemes. \textbf{Early fusion} combines sensory inputs before feeding them into shared feature extractors, where concatenation is a common way for fusion~\cite{cv4action, SEM2, PGM, Zeng2019CVPR, Carl-Lead}.
To resolve the view discrepancy, some works project point clouds on images \cite{MSFSU} or 
vice versa (predicting semantic labels for LiDAR points \cite{SDC, chen2022lav}).
On the other hand, \textbf{late fusion} combines multiple results from multi-modalities. It is less discussed due to its inferior performance \cite{xiao2020multimodal, transfuser2021CVPR}. 
Contrary to these methods, \textbf{middle fusion} achieves multi-sensor fusion within the network by separately encoding inputs and then fusing them at the feature level. Naive concatenation is also frequently adopted
\cite{MMSF, LiVi, UB-CIL, DynamicCIL, Multinet, PMP-net, Wu2022NeurIPS, jia2023thinktwice}. Recently, works have employed Transformers~\cite{transformer} to model interactions among features~~\cite{transfuser2021CVPR, transfuser2022PAMI, MMFN, Shao2022CORL, Shao2023CVPR}. 
The attention mechanism in Transformers has demonstrated great effectiveness in aggregating the context of different sensor inputs and achieving safer end-to-end driving.

Inspired by the progress in perception, it is beneficial to model modalities in a unified space such as BEV~\cite{liang2022bevfusion, liu2022bevfusion}. End-to-end driving also requires identifying policy-related contexts and discarding irrelevant details. 
\mrev{We discuss perception-based representations in Sec.~\ref{sec: challenge representation learning - perception}.}
Besides, 
the self-attention layer, interconnecting all tokens freely, incurs a significant computational cost and cannot guarantee useful information extraction. Advanced Transformer-based fusion mechanisms in the perception field, such as \cite{li2022deepfusion, borse2023xalign}, hold promise for application to the end-to-end driving task.

\subsubsection{Language as Input}
\label{sec: challenge language-guided driving}

Humans drive using both visual perception and intrinsic knowledge
which together form causal behaviors. In areas related to autonomous driving such as embodied AI, incorporating natural language as fine-grained knowledge and instructions to control the visuomotor agent has achieved notable progress~\cite{anderson2018VLN, shridhar2022cliport, duan2022survey, vemprala2023chatgpt4robotics}.
However, compared to robotic applications, \mrev{the driving task is more straightforward without the need for task decomposition}, and the outdoor environment is much more complex with highly dynamic agents but few distinctive anchors for grounding.

To incorporate linguistic knowledge into driving, a few datasets are proposed to benchmark outdoor grounding and visual language navigation tasks~\cite{deruyttere2019talk2car, Mirowski2018NIPS, chen2019touchdown, Schumann2021ACL}.
HAD \cite{advice} takes human-to-vehicle advice and adds a visual grounding task.
Sriram \textit{\etal.} \cite{LCSD} translate natural language instructions into high-level behaviors, while \cite{AL, LGD} directly ground the texts.
CLIP-MC \cite{CLIP-MC} and LM-Nav \cite{Shah2023CORL} utilize CLIP \cite{CLIP} 
to extract both linguistic knowledge from instructions and visual features from images. 

\mrev{Recently, observing the rapid development of large language models (LLMs)~\cite{openai2023gpt4, LLaMA}, works encode the perceived scene into tokens and prompt them to LLMs for control prediction and text-based explanations~\cite{Mao2023gptdriver, Xu2023DriveGPT4, Shao2024LMDrive}. Researchers also formulate the driving task as a question-answering problem and construct corresponding benchmarks~\cite{Sima2023DriveLM, Qian2024nuscenesqa}. They highlight that LLMs offer opportunities to handle sophisticated instructions and generalize to different data domains, which share similar advantages to applications in robotic areas~\cite{Yang2023survey}.}
However, LLMs for on-road driving could be challenging at present, considering their long inference time, low quantitative accuracy, and instability of outputs. 
\mrev{Potential resolutions could be employing LLMs on the cloud specifically for complex scenarios and using them solely for high-level behavior prediction.}

\subsection{\mrev{Dependence on Visual Abstraction}}
\label{sec: challenge representation learning}

End-to-end autonomous driving systems roughly have two stages: encoding the state into a latent feature representation, and then decoding the driving policy with intermediate features.
In urban driving, the input state, \textit{\ie}, the surrounding environment and ego state, is much more diverse and high-dimensional compared to common policy learning benchmarks such as video games \cite{MaRLn, ILRLSafety}, \mrev{which might lead to the misalignment between representations and necessary attention areas for policy making}.
Hence, it is helpful to \mrev{design ``good'' intermediate perception representations, or} first pre-train visual encoders using proxy tasks. This enables the network to extract useful information for driving effectively, thus facilitating the subsequent policy stage. Furthermore, this can improve the sample efficiency for RL methods.

\subsubsection{\mrev{Representation Design}}
\label{sec: challenge representation learning - perception}

\mrev{Naive representations are extracted with various backbones. Classic convolutional neural networks (CNNs) still dominate, with advantages in translation equivariance and high efficiency~\cite{He2016ResNet}. Depth-pre-trained CNNs~\cite{Lee2019VoVNet} significantly boost perception and downstream performance. In contrast, Transformer-based feature extractors~\cite{Dosovitskiy2021ViT, Dehghani2023ViT22B} show great scalability in perception tasks while not being widely adopted for end-to-end driving yet.
For driving-specific representations, researchers introduce the concept of bird's-eye-view (BEV), fusing different sensor modalities and temporal information within a unified 3D space~\cite{Li2023Delving, Li2022BEVFormer, liu2022bevfusion}. It also facilitates easy adaptions to downstream tasks~\cite{uniad, jia2023thinktwice, jia2023driveadapter, Jiang2023VAD}. In addition, grid-based 3D occupancy is developed to capture irregular objects and used for collision avoidance in planning~\cite{Tong2023OccNet}. Nevertheless, the dense representation brings huge computation costs compared to BEV methods.

Another unsettled problem is representations of the map. Traditional autonomous driving relies on HD Maps.
Due to the high cost of availability of HD Maps, online mapping methods have been devised with different formulations, such as BEV segmentation~\cite{Li2022HDMapNet}, vectorized lanlines~\cite{Liao2023MapTR}, centerlines and their topology~\cite{Wang2024OpenlaneV2, Li2023TopoNet}, and lane segments~\cite{Li2024LaneSegNet}. However, the most suitable formulation for end-to-end systems remains unvalidated.

Though various representation designs offer possibilities of how to design the subsequent decision-making process, they also place challenges as co-designing both parts is necessary for a whole framework. Besides, given the trends observed in several simple yet effective approaches with scaling up training resources~\cite{Wu2022NeurIPS, transfuser2022PAMI}, the ultimate necessity of explicit representations such as maps is uncertain. 

}

\subsubsection{\mrev{Representation Learning}}
\label{sec: challenge representation learning - pretrain}

Representation learning often incorporates certain inductive biases or prior information.
\mrev{There inevitably exist possible information bottlenecks in the learned representation, and redundant context unrelated to decisions may be removed.}

Some early methods directly utilize semantic segmentation masks from off-the-shelf networks as the input representation for subsequent policy training \cite{wang2021versatile, Behl2020IROS}. SESR \cite{SESR} further encodes segmentation masks into class-disentangled representations through a VAE \cite{VAE}. In \cite{Ahmed2022TITS, Sauer2018CORL}, predicted affordance indicators, such as traffic light states, offset to the lane center, and distance to the leading vehicle, are used as representations for policy learning. 

Observing that results like segmentation as representations can create bottlenecks defined by humans and result in loss of useful information, some have chosen intermediate features from pre-training tasks as effective representations
for RL training~\cite{MaRLn, GRI, HSMRL, DeRL}. 
In \cite{LAA}, latent features in VAE are augmented by attention maps obtained from the diffused boundary of segmentation and depth maps to highlight important regions. TARP \cite{TARP} utilizes data from a series of previous tasks to perform different tasks-related prediction tasks to acquire useful representations. In \cite{AMBS}, the latent representation is learned by approximating the $\pi$-bisimulation metric, which is comprised of differences of rewards and outputs from the dynamics model. 
ACO \cite{ACO} learns discriminative features by adding steering angle categorization into the contrastive learning structure. 
Recently, PPGeo \cite{ppgeo} proposes to learn effective representation through motion prediction together with depth estimation in a self-supervised way on uncalibrated driving videos. ViDAR \cite{Yang2024ViDAR} utilizes the raw image-point cloud pairs and pretrains the visual encoder with a point cloud forecasting pre-task.
\mrev{These works demonstrate that self-supervised representation learning from large-scale unlabeled data for policy learning is promising and worthy of future exploration.
}

\subsection{\mrev{Complexity of World Modeling for Model-based RL}}
\label{sec: challenge world model}

Besides the ability to better abstract perceptual representations, it is essential for end-to-end models to make reasonable predictions about the future to take safe maneuvers. In this section, we mainly discuss the challenges of current model-based policy learning works, where a world model provides explicit future predictions for the policy model.

Deep RL typically suffers from the high sample complexity, which is pronounced in autonomous driving.
Model-based reinforcement learning (MBRL) offers a promising direction to improve sample efficiency by allowing agents to interact with the learned world model instead of the actual environment. MBRL methods employ an explicit world (environment) model, which is composed of transition dynamics and reward functions.
This is particularly helpful in driving, as simulators like CARLA are relatively slow.

However, modeling the highly dynamic environment is a challenging task. To simplify the problem, Chen \textit{\etal.} \cite{WoR} 
factor the transition dynamics into a non-reactive world model and a simple kinematic bicycle model. 
In \cite{PGM}, a probabilistic sequential latent model is used as the world model.
To address the potential inaccuracy of the learned world model, Henaff \textit{\etal.}~\cite{Henaff2019ICLR} train the policy network with dropout regularization to estimate the uncertainty cost. Another approach~\cite{UA-MBRL} uses an ensemble of multiple world models to provide uncertainty estimation, based on which imaginary rollouts 
could be truncated and adjusted accordingly. Motivated by 
Dreamer \cite{dreamer}, ISO-Dream \cite{ISO-Dream} 
decouples visual dynamics into controllable and uncontrollable states, and trains the policy on the disentangled states.

It is worth noting that learning world models in raw image space is non-trivial for autonomous driving. Important small details, such as traffic lights, would easily be missed in predicted images. To tackle this, a few works~\cite{Yang2024GenAD, Gao2024Vista, Wang2024DriveWM} employ the prevailing diffusion technique~\cite{Rombach2022LDM}. MILE~\cite{mile2022} incorporates the Dreamer-style world model learning in the BEV segmentation space as an auxiliary task besides imitation learning. 
SEM2~\cite{SEM2} also extends the Dreamer structure but with BEV map inputs, and uses RL for training.  Besides directly using the learned world model for MBRL, DeRL~\cite{DeRL} combines a model-free actor-critic framework with the world model,
by fusing self-assessments of the action or state from both models.

World model learning for end-to-end autonomous driving is an emerging and promising direction as it greatly reduces the sample complexity for RL, and understanding the world is helpful for driving. However, as the driving environment is highly complex and dynamic, further study is still needed to determine what needs to be modeled and how to model the world effectively. 

\subsection{\mrev{Reliance on Multi-Task Learning}}
\label{sec: challenge MTL}

Multi-task learning (MTL) involves jointly performing several related tasks based on a shared representation through separate heads. MTL provides advantages such as computational cost reduction, the sharing of relevant domain knowledge, and the ability to exploit task relationships to improve model's generalization ability~\cite{caruana1997mtl}. 
Consequently, MTL is well-suited for end-to-end driving, where the ultimate policy prediction requires a comprehensive understanding of the environment.
\mrev{However, the optimal combination of auxiliary tasks and appropriate weighting of losses to achieve the best performance presents a significant challenge. }

In contrast to common vision tasks where dense predictions are closely correlated, end-to-end driving predicts a sparse signal. The sparse supervision increases the difficulty of extracting useful information for decision-making in the encoder. For image input, auxiliary tasks such as semantic segmentation~\cite{Jaeger2023ICCV,transfuser2022PAMI, MSFSU, CIL-MT, MT, FCN-LSTM} and depth estimation~\cite{Jaeger2023ICCV,transfuser2022PAMI, CIL-MT, MT, FCN-LSTM} are commonly adopted in end-to-end autonomous driving models. Semantic segmentation helps the model gain a high-level understanding of the scene;
depth estimation enables the model to capture the 3D geometry of the environment and better estimate distances to critical objects. 
Besides auxiliary tasks on perspective images, 3D object detection \cite{Jaeger2023ICCV,transfuser2022PAMI, chen2022lav} is also useful for LiDAR encoders. As BEV becomes a natural and popular representation for autonomous driving, tasks such as BEV segmentation are included in models \cite{Jaeger2023ICCV,transfuser2022PAMI, NEAT, chen2022lav, Shao2022CORL, jia2023thinktwice, zhang2023coaching, Shao2023CVPR} that aggregate features in BEV space. Moreover, in addition to these vision tasks, \cite{Shao2022CORL, MT-LfD, CIL-MT} also predict visual affordances including traffic light states, 
distances to opposite lanes, \textit{etc}.
\mrev{Nonetheless, constructing large-scale datasets with multiple types of aligned and high-quality annotations is non-trivaial for real-world applications, which remain as a great concern due to current models' reliance on MTL.}

\begin{figure}[t]
    \centering
    \includegraphics[width=.8\linewidth]{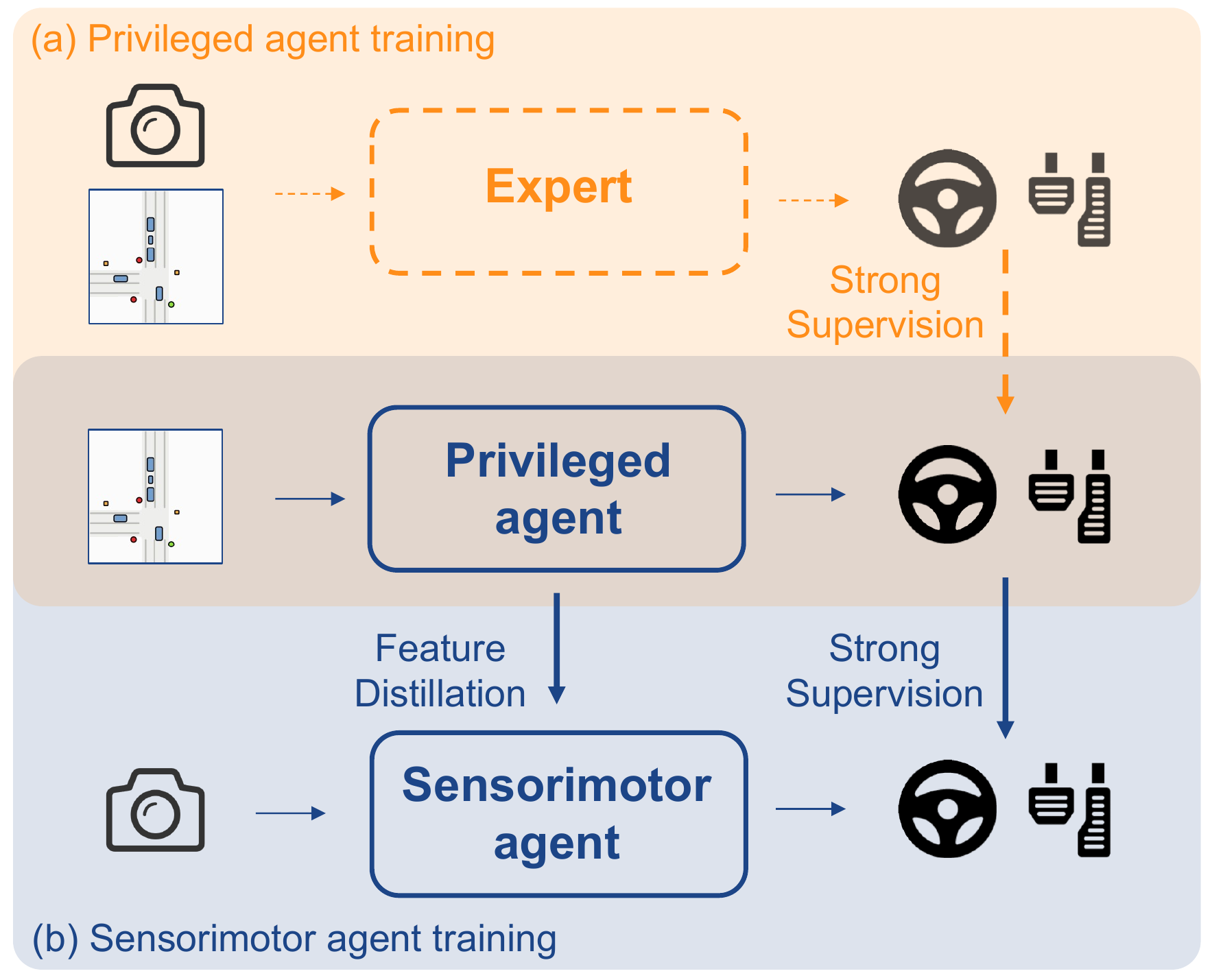}
    \caption{\textbf{Policy distillation.} (a) The \textbf{privileged agent} learns a robust policy with access to privileged ground-truth information. The expert is labeled with dashed lines to indicate that it is not mandatory if the privileged agent is trained via RL. (b) The \textbf{sensorimotor agent} imitates the privileged agent through both feature distillation and output imitation.  
    }
    \label{fig:policy-distillation}
\end{figure}

\subsection{\mrev{Inefficient Experts and Policy Distillation}}
\label{sec: challenge distillation}

As imitation learning, or its predominant sub-category, behavior cloning, is simply supervised learning that mimics expert behaviors, corresponding methods usually follow the ``Teacher-Student'' paradigm. 
\mrev{There lie two main challenges:
(1) Teachers, such as the handcrafted expert autopilot provided by CARLA, are not perfect drivers, though having access to ground-truth states of surrounding agents and maps.
(2) Students are supervised by the recorded output with sensor input only, requiring them to extract perceptual features and learn policy from scratch simultaneously.}

A few studies propose to divide the learning process into two stages, \textit{\ie}, training a stronger teacher network and then distilling the policy to the student. In particular, Chen \textit{\etal.}~\cite{LBC, chen2022lav} first employ a privileged agent to learn how to act with access to the state of the environment, then let the sensorimotor agent (student) closely imitate the privileged agent with distillation at the output stage. More compact BEV representations as input for the privileged agent provide stronger generalization abilities and supervision than the original expert. The process is depicted in Fig.~\ref{fig:policy-distillation}. 

Apart from solely supervising planning results, several works also distill knowledge at the feature level. For example, FM-Net \cite{FM-Net} employs segmentation and optical flow models as auxiliary teachers to guide feature training. SAM \cite{SAM} adds L2 feature loss between teacher and student networks,
while CaT \cite{zhang2023coaching} aligns features in BEV.
WoR \cite{WoR} learns a model-based action-value function and then uses it to supervise the visuomotor policy. 
Roach \cite{roach} trains a stronger privileged expert with RL, eliminating the upper bound of BC. It incorporates multiple distillation targets, \textit{\ie}, action distribution, values/rewards, and latent features. By leveraging the powerful RL expert, TCP \cite{Wu2022NeurIPS} achieves a new state-of-the-art on the CARLA leaderboard with a single camera as visual input. DriveAdpater~\cite{jia2023driveadapter} learns a perception-only student and adapters with the feature alignment objective. The decoupled paradigm fully enjoys the teacher's knowledge and student's training efficiency.

Though huge efforts have been devoted to designing a robust expert and transferring knowledge at various levels, the teacher-student paradigm still suffers from inefficient distillation. 
For instance, the privileged agent has access to ground-truth states of traffic lights, which are small objects in images and thus hard to distill corresponding features. As a result, the visuomotor agents exhibit large performance gaps compared to their privileged agents. It may also lead to causal confusion for students (see Sec.~\ref{sec:causal_confusion}). It is worth exploring how to draw more inspiration from general distillation methods in machine learning to minimize the gap.

\begin{figure}[t]
    \centering
    \includegraphics[width=\linewidth]{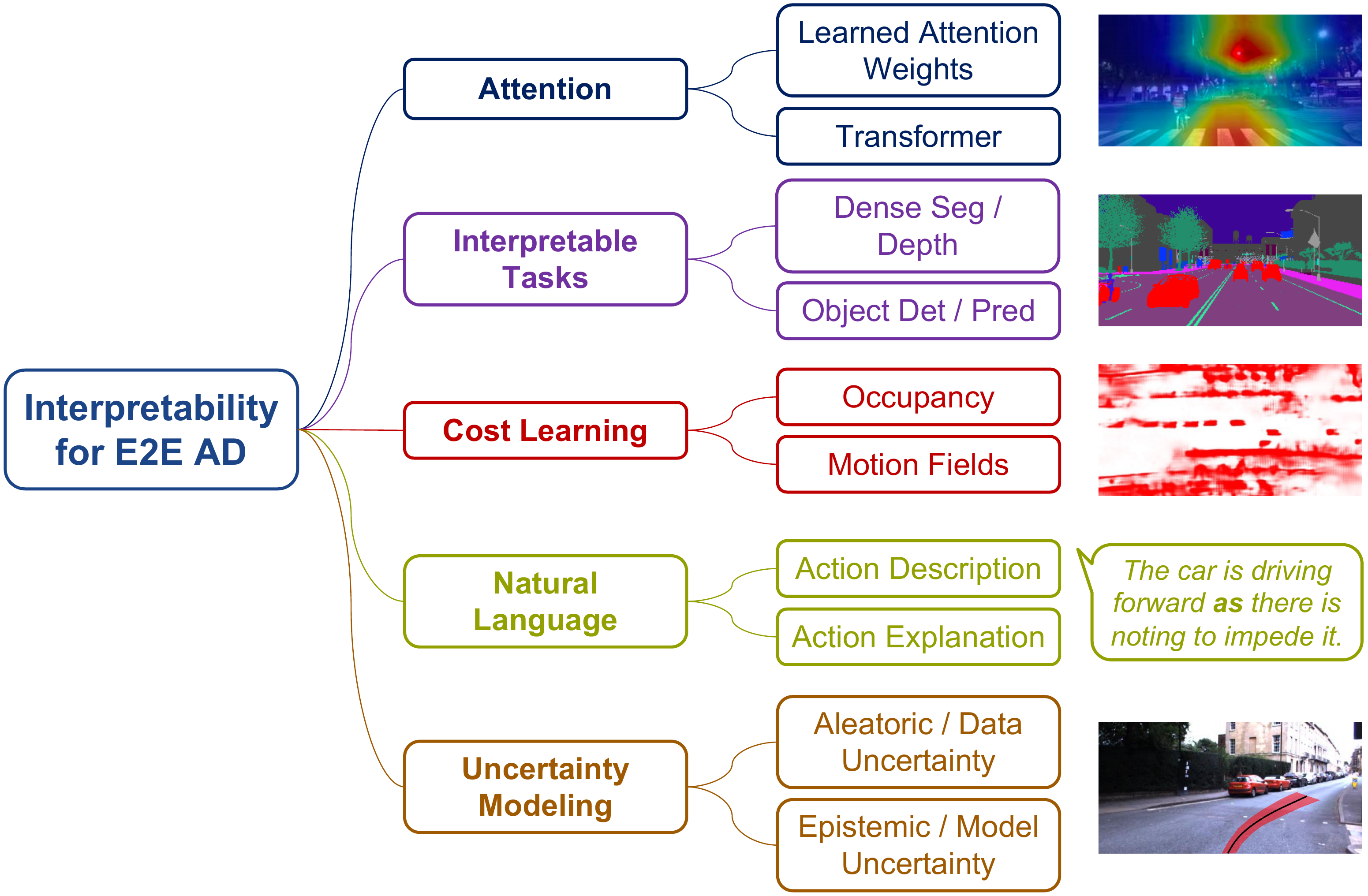}
    \caption{\textbf{Summary of the different forms of interpretability.} They aid in human comprehension of decision-making processes of end-to-end models
    and the reliability of outputs.}
    \label{fig:interpretability}
\end{figure}

\subsection{\mrev{Lack of Interpretability}}
\label{sec: challenge interpretability}

Interpretability plays a critical role in autonomous driving \cite{zablocki2022explainability}. It enables engineers to better debug the system, provides performance guarantees from a societal perspective, 
and promotes public acceptance. Achieving interpretability 
for end-to-end driving models, which are often referred to as ``black boxes'', is more essential and challenging.

Given trained models, some post-hoc X-AI (explainable AI) techniques could be applied to gain saliency maps~\cite{PilotNet, visualbackprop, mohseni2019predicting, kim2017interpretable, CIL-MT}. Saliency maps highlight specific regions in the visual input on which the model primarily relies for planning. However, this approach provides limited information, and its effectiveness and validity are difficult to evaluate. Instead, we focus on end-to-end frameworks that directly enhance interpretability in their model design. We introduce each category of interpretability in Fig.~\ref{fig:interpretability} below.

\textbf{Attention Visualization:} The attention mechanism provides a certain degree of interpretability. In \cite{kim2017interpretable, mori2019visual, CIL-MT, MT-LfD, BDD-X}, a learned attention weight is applied to aggregate important features from intermediate feature maps. Attention weights can also adaptively combine ROI pooled features from different object regions \cite{wang2019deep} or a fixed grid~\cite{IL-Att}. NEAT \cite{NEAT} iteratively aggregates features to predict attention weights and refine the aggregated feature. Recently, 
Transformer attention blocks are employed to better fuse different sensor inputs, and attention maps display important regions in the input for driving decisions~\cite{Xiao2023ARXIV, transfuser2022PAMI, Jaeger2023ICCV, Shao2022CORL, MMFN}. In PlanT \cite{PlanT}, attention layers process features from different vehicles, providing interpretable insights into the corresponding action. Similar to post-hoc saliency methods, although attention maps offer straightforward clues about models' focus, their faithfulness and utility remain limited. 

\textbf{Interpretable Tasks:} 
Many IL-based works introduce interpretability by decoding the latent feature representations into other meaningful information besides policy prediction, such as semantic segmentation \cite{uniad, Shao2022CORL, chen2022lav, NEAT, transfuser2022PAMI, Jaeger2023ICCV, CLIP-MC, CIL-MT, UB-CIL, MSFSU, MT, FCN-LSTM, Sun2023ITSC}, depth estimation \cite{transfuser2022PAMI, Jaeger2023ICCV, CIL-MT, UB-CIL, MT}, object detection \cite{uniad, chen2022lav, transfuser2022PAMI, Jaeger2023ICCV}, affordance predictions \cite{Shao2022CORL, CIL-MT, MT-LfD}, motion prediction \cite{uniad, chen2022lav}, and gaze map estimation \cite{EyeGazeIL}. Although these methods provide interpretable information, most of them only treat these predictions as auxiliary tasks \cite{transfuser2022PAMI, Jaeger2023ICCV, NEAT, CIL-MT, UB-CIL, MSFSU, MT, MT-LfD}, with no explicit impact on final driving decisions. Some \cite{Shao2022CORL, chen2022lav} do use these outputs for final actions, but they are incorporated solely for performing an additional safety check.

\textbf{\mrev{Rules Integration and Cost Learning:}} As discussed in Sec.~\ref{sec:IOC}, cost learning-based methods share similarities with traditional modular systems and thus exhibit a certain level of interpretability. NMP \cite{Zeng2019CVPR} and DSDNet \cite{DSDNet} construct the cost volume in conjunction with detection and motion prediction results. P3 \cite{P3} combines predicted semantic occupancy maps with comfort and traffic rules constraints to construct the cost function. Various representations, such as probabilistic occupancy and temporal motion fields \cite{Casas2021CVPR}, emergent occupancy \cite{EO}, and freespace \cite{FF}, are employed to score sampled trajectories. \mrev{In \cite{Lookout, hu2022stp3, Dauner2023ARXIV, Jiang2023VAD}, human expertise and pre-defined rules including safety, comfort, traffic rules, and routes based on perception and prediction outputs are explicitly included to form the cost for trajectory scoring, demonstrating improved robustness and safety.}

\textbf{Linguistic Explainability:} As one aspect of interpretability is to help humans understand the system, natural language is a suitable choice for this purpose. Kim \textit{\etal}~\cite{BDD-X} and Xu \textit{\etal}~\cite{xu2020explainable} develop datasets pairing driving videos or images with descriptions and explanations, and propose end-to-end models with both control and explanation outputs.
BEEF \cite{BEEF} 
fuses the predicted trajectory and the intermediate perception features to predict justifications for the decision. 
ADAPT \cite{ADAPT} proposes a Transformer-based network to jointly estimate action, narration, and reasoning.
\mrev{Recently, \cite{Sima2023DriveLM, Qian2024nuscenesqa, Xu2023DriveGPT4} resort to the progress of multi-modality and foundation models, using LLMs/VLMs to provide decision-related explanations, as discussed in Sec.~\ref{sec: challenge language-guided driving}.
}

\textbf{Uncertainty Modeling:} \mrev{Uncertainty is a quantitative approach for interpreting the dependability of deep learning model outputs~\cite{Guo2017ICML, Loquercio2020RAL},
which can be helpful for designers and users to identify uncertain cases for improvement or necessary intervention.}
For deep learning, there are two types of uncertainty: aleatoric uncertainty and epistemic uncertainty. Aleatoric uncertainty is inherent to the task, while epistemic uncertainty is due to limited data or modeling capacity. In \cite{uncertainty-eval}, 
authors leverage certain stochastic regularizations in the model to perform multiple forward passes as samples to measure the uncertainty. However, the requirement of multiple forward passes is not feasible in real-time scenarios. \mrev{Loquercio \textit{\etal}}~\cite{Loquercio2020RAL} and Filos \textit{\etal}~\cite{RIP} propose capturing epistemic uncertainty with an ensemble of expert likelihood models and aggregating the results to perform safe planning. Regarding methods modeling aleatoric uncertainty, driving actions/planning and uncertainty (usually represented by variance) are explicitly predicted in \cite{UAIL, VTGNet, PMP-net}. \mrev{Such methods directly model and quantify the uncertainty at the action level as a variable for the network to predict. The planner would generate the final action based on the predicted uncertainty, either choosing the action with the lowest uncertainty from multiple actions \cite{UAIL} or generating a weighted combination of proposed actions based on the uncertainties \cite{PMP-net}.} Currently, predicted uncertainty is mainly utilized in combination with hard-coded rules. Exploring better ways to model and utilize uncertainty for autonomous driving is necessary.

\subsection{\mrev{Lack of Safety Guarantees}}
\label{sec: challenge safety guarantee}

\mrev{Ensuring safety is of utmost importance when deploying autonomous driving systems in real-world scenarios. However, the learning-based nature of end-to-end frameworks inherently lacks precise mathematical guarantees regarding safety, 
unlike traditional rule-based approaches~\cite{shalev2017formal}. 

Nevertheless, it should be noted that modular driving stacks have already incorporated specific safety-related constraints or optimizations within their motion planning or speed prediction modules to enforce safety~\cite{brudigam2021stochastic, lyu2021probabilistic, allamaa2024real}.
These mechanisms can potentially be adapted for integration into end-to-end models as post-process steps or safety checks, thereby providing additional safety guarantees. Furthermore, the intermediate interpretability predictions, as discussed in Sec.~\ref{sec: challenge interpretability}, such as detection and motion prediction results, can be utilized in post-processing procedures.

}

\begin{figure}[t]
    \centering
    \includegraphics[width=.9\linewidth]{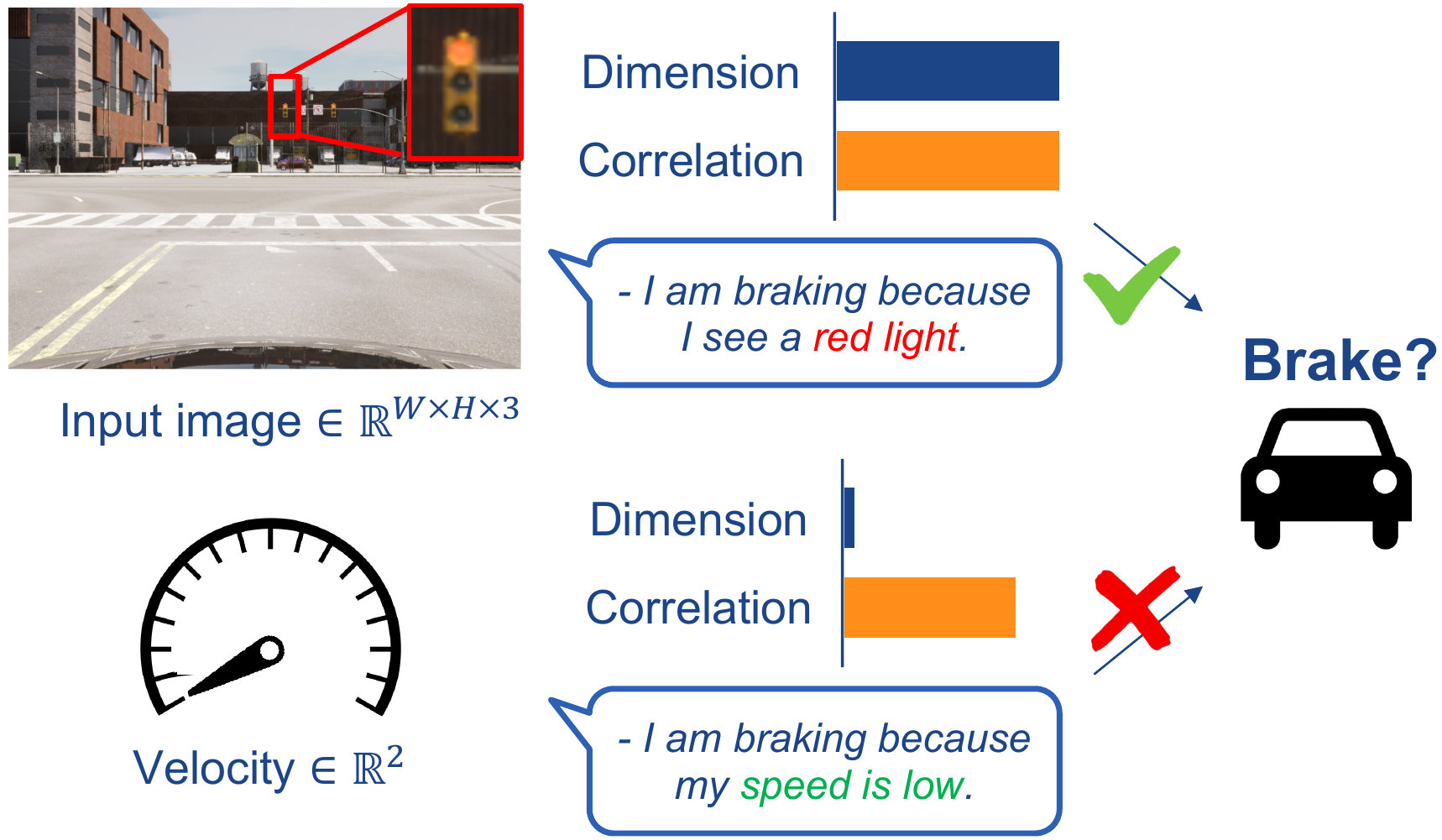}
    \caption{\textbf{Causal Confusion.} The current action of a car is strongly correlated with low-dimensional spurious features such as the velocity or the car's past trajectory. End-to-End models may latch on to them leading to causal confusion.}
    \label{fig:causal-confusion}
\end{figure}

\begin{figure*}[t]
    \centering
    \includegraphics[width=.95\linewidth]{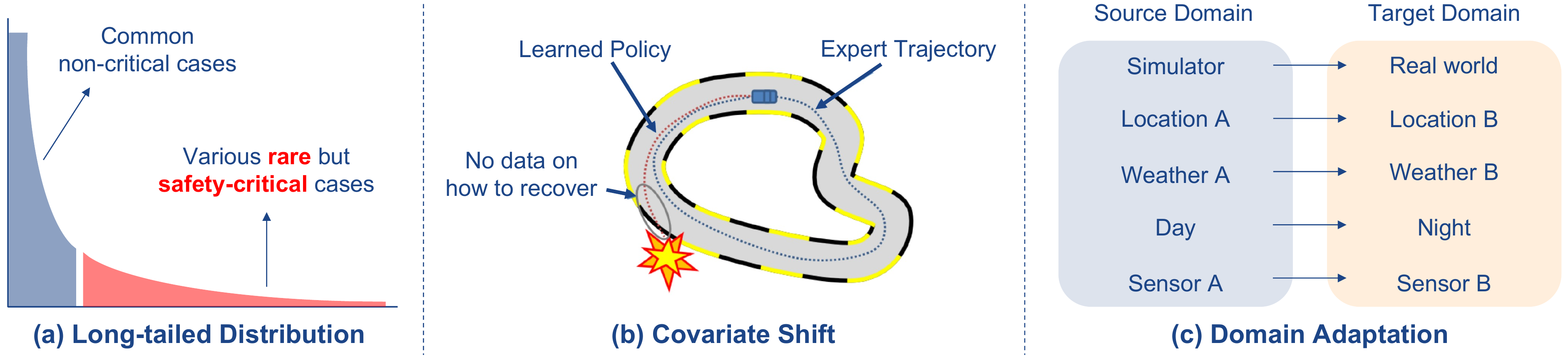}
    \caption{\textbf{Challenges in robustness.} Three primary generalization issues arise in relation to dataset distribution discrepancies, namely long-tailed and normal cases, expert demonstration and test scenarios, and domain shift in locations, weather, \textit{etc}.}
    \label{fig:generalization}
\end{figure*}

\subsection{Causal Confusion}
\label{sec:causal_confusion}

Driving is a task that exhibits temporal smoothness, which makes past motion a reliable predictor of the next action. However, methods trained with multiple frames can become overly reliant on this shortcut \cite{Geirhos2020NatureMI} and suffer from catastrophic failure during deployment.
This problem is referred to as the copycat problem \cite{Wen2020NEURIPS} in some works and is a manifestation of causal confusion \cite{Haan20219NEURIPS}, where access to more information leads to worse performance. 

Causal confusion in imitation learning has been a persistent challenge for nearly two decades.
One of the earliest reports of this effect was made by LeCun \textit{\etal} \cite{Muller2005NEURIPS}. They used a single input frame for steering prediction to avoid such extrapolation. Though simplistic, this is still a preferred solution in current state-of-the-art IL methods \cite{transfuser2022PAMI, Wu2022NeurIPS}. Unfortunately, using a single frame makes it hard to extract the motion of surrounding actors. Another source of causal confusion is speed measurement \cite{CILRS}. Fig.~\ref{fig:causal-confusion} showcases an example of a car waiting at a red light. The action of the car could highly correlate with its speed because it has waited for many frames where the speed is zero and the action is the brake. Only when the traffic light changes from red to green does this correlation break down.

There are several approaches to combat the causal confusion problem when using multiple frames. 
In \cite{Wen2020NEURIPS}, the authors attempt to remove spurious temporal correlations from the bottleneck representation by training an adversarial model that predicts the ego agent's past action. 
Intuitively, the resulting min-max optimization trains the network to eliminate its past from intermediate layers.
It works well in MuJoCo but does not scale to complex vision-based driving. 
OREO \cite{Park2021NEURIPS} maps images to discrete codes representing semantic objects and applies random dropout masks to units that share the same discrete code,
which helps in confounded Atari.
In end-to-end driving, ChauffeurNet \cite{Bansal2019RSS} addresses the causal confusion issue by using the past ego-motion as intermediate BEV abstractions and dropping out it with a 50\% probability during training.
Wen \textit{\etal.}~\cite{Wen2021ICML} propose upweighting keyframes in the training loss, where a decision change occurs (and hence are not predictable by extrapolating the past). 
PrimeNet \cite{Wen2022ICML} improves performance compared to keyframes by using an ensemble, where the prediction of a single-frame model is given as additional input to a multi-frame model. Chuang \textit{\etal.} \cite{Chuang2022ECCV} do the same but supervise the multi-frame network with action residuals instead of actions. 
In addition, the problem of causal confusion can be circumvented by using only LiDAR histories (with a single frame image) and realigning point clouds into one coordinate system. This removes ego-motion while retaining information about other vehicles' past states. This technique has been used in multiple works \cite{Zeng2019CVPR, Casas2021CVPR, chen2022lav}, though it was not motivated this way.

However, these studies have used environments that are modified to simplify studying the causal confusion problem. Showing performance improvements in state-of-the-art settings as mentioned in Sec.~\ref{sec:benchmarks} remains an open problem.

\subsection{\mrev{Lack of Robustness}}
\label{sec: challenge generalizability}

\subsubsection{Long-tailed Distribution} 
\label{sec: long-tail}
One important aspect of the long-tailed distribution problem is dataset imbalance, where a few classes make up the majority, 
as shown in Fig.~\ref{fig:generalization} (a). This poses a big challenge for models to generalize to diverse environments. Various methods mitigate this issue with data processing, including over-sampling~\cite{buda2018systematic, byrd2019effect}, under-sampling~\cite{mani2003knn, liu2008exploratory}, and data augmentation~\cite{gidaris2018dynamic, zhang2017mixup}. Besides, weighting-based approaches \cite{lin2017focal, cui2019class} are also commonly used.

In the context of end-to-end autonomous driving, the long-tailed distribution issue is particularly severe. 
Most drives are repetitive and uninteresting \textit{\eg}, following a lane for many frames. \mrev{Conversely, interesting safety-critical scenarios occur rarely but are diverse in nature, and hard to replicate in the real world for safety reasons.} To tackle this, some works rely on handcrafted scenarios \cite{ETL, carlaleaderboard, PGDrive, SUMO, Suo2021CVPR} to generate more diverse data in simulation. LBC \cite{LBC} leverages the privileged agent to create imaginary supervisions conditioned on different navigational commands. LAV \cite{chen2022lav}
includes trajectories of non-ego agents for training to promote data diversity. In \cite{o2018scalable}, a simulation framework is proposed to apply importance-sampling strategies to accelerate the evaluation of rare-event probabilities.

Another line of research \cite{2019generating, Learningtocollide, ding2021multimodal, Advsim, KING, zhang2023cat} generates safety-critical scenarios in a data-driven manner through adversarial attacks. In \cite{2019generating}, Bayesian Optimization is employed to generate adversarial scenarios. Learning to collide \cite{Learningtocollide} represents driving scenarios as the joint distribution over building blocks and applies policy gradient RL methods to generate risky scenarios. AdvSim \cite{Advsim} modifies agents' trajectories to cause failures, while still adhering to physical plausibility.
KING \cite{KING} proposes an optimization algorithm for safety-critical perturbations using gradients through differentiable kinematics models.

In general, efficiently generating realistic safety-critical scenarios that cover the long-tailed distribution remains a significant challenge. While many works focus on adversarial scenarios in simulators, it is also essential to better utilize real-world data for critical scenario mining and potential adaptation to simulation. Besides, a systematic, rigorous, comprehensive, and realistic testing framework is crucial for evaluating end-to-end autonomous driving methods under these long-tailed distributed safety-critical scenarios. 

\subsubsection{Covariate Shift}
\label{sec: covariate-shift}

As discussed in Sec.~\ref{sec:IL}, one important challenge for behavior cloning is covariate shift. The state distributions from the expert's policy and those from the trained agent's policy differ, leading to compounding errors when the trained agent is deployed in unseen testing environments or when the reactions from other agents differ from training time. This could result in the trained agent being in a state that is outside the expert's distribution for training, leading to severe failures. An illustration is presented in Fig.~\ref{fig:generalization} (b). 

DAgger (Dataset Aggregation) \cite{DAgger} is a common solution for this issue. DAgger is an iterative training process. The current trained policy is rolled out in each iteration to collect new data, and the expert is used to label the visited states. This enriches the dataset by adding examples of how to recover from suboptimal states that an imperfect policy might visit. The policy is then trained on the augmented dataset, and the process repeats. However, one downside of DAgger is the need for an available expert to query online.

For end-to-end autonomous driving, DAgger is adopted in \cite{AgileAD} with an MPC-based expert. To reduce the cost of constantly querying the expert,
SafeDAgger \cite{SafeDAgger} extends the original DAgger algorithm by learning a safety policy that estimates the deviation between the current policy and the expert policy. The expert is only queried when the deviation is large.
MetaDAgger \cite{MetaDAgger} uses meta-learning with DAgger to aggregate data from multiple environments. LBC \cite{LBC} adopts DAgger and resamples the data with higher loss more frequently. 
In DARB \cite{DARB}, 
to better utilize failure or safety-related samples, it proposes several mechanisms, including task-based, policy-based, and policy \& expert-based mechanisms, to sample such critical states.

\subsubsection{Domain Adaptation} \label{sec: domain-adaptation}
Domain adaptation (DA) is a type of transfer learning in which the target task is the same as the source task, but the domains differ. Here we discuss scenarios where labels are available for the source domain while there are no labels or a limited amount of labels available for the target domain. 

As shown in Fig.~\ref{fig:generalization} (c), domain adaptation for autonomous driving tasks encompasses several cases \cite{DA2021survey}: 
\begin{itemize}
    \item Sim-to-real: the large gap between simulators used for training and the real world used for deployment.
    \item Geography-to-geography: different geographic locations with varying environmental appearances.
    \item Weather-to-weather: changes in sensor inputs caused by weather conditions such as rain, fog, and snow.
    \item Day-to-night: illumination variations in visual inputs.
    \item Sensor-to-sensor: possible differences in sensor characteristics, \textit{\eg}, resolution and relative position.
\end{itemize}
Note that the aforementioned cases often overlap. 

Typically, domain-invariant feature learning is achieved with image translators and discriminators to map images from two domains into a common latent space or representations like segmentation maps~\cite{VISRI, sim2real}. 
LUSR~\cite{LUSR} and UAIL~\cite{UAIL} adopt a Cycle-Consistent VAE and GAN, respectively, to project images into a latent representation comprised of a domain-specific part and a domain-general part.
In SESR \cite{SESR}, class disentangled encodings are extracted from a semantic segmentation mask to reduce the sim-to-real gap.
Domain randomization \cite{tobin2017domain, peng2018sim, matas2018sim} is also a simple and effective sim-to-real technique for RL policy learning, which is further adapted for end-to-end autonomous driving \cite{carla-ppo, wang2021versatile}. It is realized by randomizing the rendering and physical settings of the simulators to cover the variability of the real world during training.

Currently, sim-to-real adaptation through source target image mapping or domain-invariant feature learning is the focus.
Other DA cases 
are handled by constructing a diverse and large-scale dataset. Given that current methods mainly concentrate on the visual gap in images, and LiDAR has become a popular input modality for driving, specific adaptation techniques tailored for LiDARs must also be designed. Besides, traffic agents' behavior gaps between the simulator and the real world should be noticed as well. Incorporating real-world data into simulation through techniques such as NeRF \cite{mildenhall2020nerf} is another promising direction.

\section{Future Trends}
\label{sec: future trends}
Considering the challenges and opportunities discussed, we list some crucial directions for future research that may have a broader impact in this field.

\subsection{Zero-shot and Few-shot Learning}
\label{sec: future zero shot}

It is inevitable for autonomous driving models to eventually encounter real-world scenarios that lie beyond the training data distribution. This raises the question of whether we can successfully adapt the model to an unseen target domain where limited or no labeled data is available. Formalizing this task for the end-to-end driving domain and incorporating techniques from the zero-shot/few-shot learning literature are the key steps toward achieving this~\cite{Kirk2023JAIR, snell2017prototypical}.

\subsection{Modular End-to-end Planning}
\label{sec: future modular E2E}

The modular end-to-end planning framework optimizes multiple modules while prioritizing the ultimate planning task, which enjoys the advantages of interpretability as indicated in Sec.~\ref{sec: challenge interpretability}. This is advocated in recent literature \cite{Karkus2022CORL, uniad} and certain industry solutions (Tesla, Wayve, \textit{\etc}) have involved similar ideas. When designing these differentiable perception modules, several questions arise regarding the choice of loss functions, such as the necessity of 3D bounding boxes for object detection, 
whether opting for BEV segmentation over lane topology for static scene perception, or the training strategies with limited modules' data.

\subsection{Data Engine}
\label{sec: future data engine}

The importance of large-scale and high-quality data for autonomous driving can never be emphasized enough~\cite{li2023open}. Establishing a data engine with an automatic labeling pipeline \cite{kirillov2023segany} could greatly facilitate the iterative development of both data and models. The data engine for autonomous driving, especially modular end-to-end planning systems, needs to streamline the process of annotating high-quality perception labels with the aid of large perception models in an automatic way. It should also support mining hard/corner cases, scene generation, and editing to facilitate the data-driven evaluations discussed in Sec.~\ref{sec: Online Evaluation} and promote diversity of data and the generalization ability of models (Sec.~\ref{sec: challenge generalizability}). A data engine would enable autonomous driving models to make consistent improvements.

\subsection{Foundation Model}
\label{sec: future foundation model}

Recent advancements in foundation models in both language~\cite{openai2023gpt4, LLaMA} and vision~\cite{kirillov2023segany, DINOv2} have 
proved that large-scale data and model capacity can unleash the immense potential of AI in high-level reasoning tasks. The paradigm of finetuning~\cite{flamingo} or prompt learning~\cite{ouyang2022instructgpt}, optimization in the form of self-supervised reconstruction~\cite{MAE} or contrastive pairs~\cite{CLIP}, \textit{\etc}, are all applicable to the end-to-end driving domain.
However, we contend that the direct adoption of LLMs for driving might be tricky.
The output of an autonomous agent requires steady and accurate measurements, whereas the generative output in language models aims to behave like humans, irrespective of its accuracy. 
A feasible solution to develop a ``foundation'' driving model is to train a world model that can forecast the reasonable future of the environment, either in 2D, 3D, or latent space. To perform well on downstream tasks like planning, the objective to be optimized for the model needs to be sophisticated enough, beyond frame-level perception.

\section{\mrev{Conclusion and Outlook}}
\label{sec:conclusion}

In this survey, we provide an overview of fundamental methodologies and summarize various aspects of simulation and benchmarking. We thoroughly analyze the extensive literature to date, and highlight a wide range of critical challenges and promising resolutions. %

\smallskip
\mrev{
\textbf{Outlook:}
The industry has dedicated considerable effort over the years to develop advanced modular-based systems capable of achieving self-driving on highways. However, these systems face significant challenges when confronted with complex scenarios, \textit{\eg}, inner-city streets and intersections. Therefore, an increasing number of companies have started exploring end-to-end autonomous driving techniques specifically tailored for these environments. It is envisioned that with extensive high-quality data collection, large-scale model training, and the establishment of reliable benchmarks, 
the end-to-end approach will have enormous potential over modular stacks in terms of performance and effectiveness.
In summary,
}
end-to-end autonomous driving faces great opportunities and challenges simultaneously, with the ultimate goal of building generalist agents. In this era of emerging technologies, we hope this survey could serve as a starting point to shed new light on this domain.

\section*{Acknowledgments}

This project is partially supported by National Key R\&D Program of China (2022ZD0160104), NSFC (62206172), and Shanghai Committee of Science and Technology (23YF1462000). 
Andreas Geiger and Bernhard Jaeger are supported by the ERC Starting Grant LEGO-3D (850533), the BMWi in the project KI Delta Learning (project number 19A19013O), and the DFG EXC number 2064/1 - project number 390727645. Kashyap Chitta is supported by the German Federal Ministry of Education and Research (BMBF): Tübingen AI Center, FKZ: 01IS18039A. We thank the International Max Planck Research School for Intelligent Systems for supporting Bernhard Jaeger and Kashyap Chitta.

\bibliographystyle{ieeetr}
\bibliography{bibliography_short, bibliography, bibliography_custom}

\end{document}